\pdfoutput=1

\documentclass[11pt]{article}

\usepackage[preprint]{acl}

\usepackage{times}
\usepackage{latexsym}
\usepackage{makecell} 
\usepackage{booktabs}
\usepackage{caption}
\usepackage{subcaption}
\usepackage{enumitem}
\usepackage{comment}
\usepackage{amsmath}

\usepackage[T1]{fontenc}

\usepackage[utf8]{inputenc}

\usepackage{microtype}

\usepackage{inconsolata}

\usepackage{graphicx}

\title{Can Third-parties Read Our Emotions?}

\author{
  Jiayi Li, Yingfan Zhou, Pranav Narayanan Venkit, Halima Binte Islam, Sneha Arya \\   \textbf{Shomir Wilson, Sarah Rajtmajer} \\
  Pennsylvania State University \\
  \texttt{\{jpl6207, yxz5975, pranav.venkit, hpi5039, sfa5848, shomir, smr48\}@psu.edu}
}

\begin{document}
\maketitle
\begin{abstract} Natural Language Processing tasks that aim to infer an author's private states, e.g., emotions and opinions, from their written text, typically rely on datasets annotated by third-party annotators. However, the assumption that third-party annotators can accurately capture authors' private states remains largely unexamined. In this study, we present human subjects experiments on emotion recognition tasks that directly compare third-party annotations with first-party (author-provided) emotion labels. Our findings reveal significant limitations in third-party annotations—whether provided by human annotators or large language models (LLMs)—in faithfully representing authors' private states. However, LLMs outperform human annotators nearly across the board. We further explore methods to improve third-party annotation quality. We find that demographic similarity between first-party authors and third-party human annotators enhances annotation performance. While incorporating first-party demographic information into prompts leads to a marginal but statistically significant improvement in LLMs' performance. We introduce a framework for evaluating the limitations of third-party annotations and call for refined annotation practices to accurately represent and model authors' private states.

\end{abstract}

\section{Introduction}
Recognizing and interpreting subjective language used to express \emph{private states}--users' internal experiences, e.g., opinions, emotions, evaluations, and speculations \cite{quirk1985grammar}--has been a long-standing area of study in natural language processing (NLP) \cite{wiebe2004learning, Banfield1982UnspeakableS}. 
Examples of private state inference tasks include emotion classification \cite{mohammad-etal-2018-semeval,demszky-etal-2020-goemotions}, emotion intensity detection \cite{mohammad-bravo-marquez-2017-emotion}, sarcasm detection \cite{bamman2015contextualized, oprea2019isarcasm}, sentiment analysis \cite{nemes2021social}, 
political ideology detection \cite{iyyer2014political}, stance detection \cite{aldayel2021stance}, and many more. Approaches for these tasks have relied heavily on `gold standard' datasets annotated by third-party annotators. The standard annotation process typically involves multiple human annotators, often recruited through crowdsourcing platforms or drawn from internal research teams \cite{rashtchian2010collecting, snow2008cheap}. More recently, large language models (LLMs) have been explored as scalable and cost-effective alternatives or complements to human annotators \cite{li2023coannotating}. Studies have shown that LLMs can match or surpass human annotators on some annotation tasks \cite{DBLP:conf/emnlp/HasanainAA24, DBLP:journals/corr/abs-2303-15056}.

While the adoption of third-party annotations is appropriate for many NLP tasks such as those with objective ground truth (e.g., named entity recognition \cite{nadeau2009survey}, part-of-speech tagging \cite{brill1994some}) or for tasks that benefit from diverse third-party perception (e.g., toxicity detection \cite{DBLP:conf/acl/PavlopoulosSDTA20}, hate speech detection \cite{davidson2017automated}), \textbf{\emph{we suggest that third-party annotations (human or machine) have inherent limitations for tasks that seek to model an author's private state}}. Fundamentally, the use of third-party annotations assumes that an author’s private state can be identified from their writing by human or machine annotators. However, subjective language often lacks explicit linguistic cues \cite{wiebe2005annotating, balahur2012detecting}, leaving annotators to perform inference on textual cues, which may be implicit, ambiguous, or context-dependent. This challenge is compounded by individual differences in authors' expression of private states \cite{deandrea2010online, bauer2003representation} as well as annotators' socio-demographic, cultural backgrounds, and personal beliefs which shape how they interpret authors' text \cite{shen-rose-2021-sounds, ding2022impact, oprea2019isarcasm}. 


Misalignment between an author’s private state and its interpretation by third-party annotators is not merely a labeling error—it can propagate through learned models and compromise the reliability of downstream applications. 
However, little research has systematically examined this gap. 

To this end, we conduct a series of studies to investigate the alignment between third-party annotations and first-party labels, i.e., authors' self-reported private states, in the context of emotion recognition tasks. 
Emotion recognition has a wide range of applications, especially in high-stakes contexts such as moderating online content, detecting deception, and powering therapeutic chatbots \cite{shaw2017lie, chiril2022emotionally, zygadlo2021text}. In these contexts, misinterpretations of users' emotions not only render the technology deficient but also make it socially pernicious. Moreover, emotion recognition is closely related to other tasks that aim to infer private states, such as mental health detection and sentiment analysis \cite{zhang2023emotion, venkit-etal-2023-sentiment} so that we anticipate lessons learned in this domain will inform the next steps in others.  

\noindent Our work is guided by the research questions:

\noindent\textbf{RQ1}: How do third-party human annotators and LLMs align with first-party labels, i.e., authors' self-reports, on emotion recognition tasks? 

\noindent\textbf{RQ2}: Does demographic similarity between human annotators and authors improve alignment between third-party annotations and first-party labels?

\noindent\textbf{RQ3}: Does including authors' demographic information within prompts improve LLM annotations?

To explore these questions, we recruit social media users to share their own social media posts and label them with their own emotion labels (first party). We then collect annotations of these same posts from third-party human annotators and LLMs and compare their performance. 
We investigate the impact of demographic similarity between third-party human annotators and first-party authors; similarly, we explore whether including first-party demographic information within prompts improves LLM annotations.

Our results reveal notable misalignment between first-party labels and third-party annotations, as both human annotators and LLMs achieve low to fair Cohen's kappa scores and F1 scores across emotions. LLMs perform better than human annotators universally \textbf{(RQ1)}. 
In-group human annotators' performance is significantly better than out-group annotators, suggesting that demographic similarity between third-party human annotators and first-party individuals improves annotation performance (\textbf{RQ2}). We further observe that adding first-party information in prompts marginally improves LLMs' annotations (\textbf{RQ3}).

\section{Related Work}
\subsection{Private State Annotations for NLP} Emotions, sentiment, opinions, evaluations, and speculations are fundamental aspects of an individual's internal world, collectively referred to as private states \cite{quirk1985grammar}. 
Many NLP tasks aim to identify an author's private states from their texts. Labeled datasets are needed to train and test models to perform these tasks. 

\noindent \textbf{Human-annotated datasets.} For emotion recognition, \cite{mohammad-etal-2018-semeval} introduce a set of subtasks, such as emotion classification and emotion intensity classification, that aim to infer the effectual state of a person from their tweets and provide a human-annotated dataset for each of the subtasks. \cite{aman2007identifying} introduce a corpus that consists of blog posts with human annotations of emotion on sentence level. \cite{demszky-etal-2020-goemotions} build a dataset of Reddit posts labeled by third-party human annotators based on a fine-grained emotion taxonomy. These datasets have been widely adopted for model building. 

\noindent \textbf{LLM-annotated datasets. }Given LLMs' demonstrated zero-shot and few-shot capabilities for various NLP tasks \cite{brown2020language}, researchers have increasingly explored their use to augment or replace human annotators \cite{DBLP:journals/corr/abs-2303-15056, kim-etal-2024-meganno}. Studies have highlighted the potential of LLMs as annotators even for subjective tasks that model authors' private states, such as emotion recognition \cite{DBLP:journals/corr/abs-2408-17026} and stance detection \cite{li2024advancing}. 

\noindent \textbf{Limitations of third-party annotations.} 
Limited research has systematically investigated the limitations of third-party annotations for private states. Oprea et al. \cite{oprea2019isarcasm} examined the differences between intended sarcasm (as labeled by the author) and perceived sarcasm (as labeled by third-party annotators). While sarcasm detection can benefit from explicit linguistic cues, such as irony and exaggeration, understanding private states often relies on implicit signals and is subject to individual interpretation.

Another relevant study \cite{joseph-etal-2021-mis} explored the misalignment between third-party inferences of users' political stances and users' responses in public opinion surveys. Rather than directly comparing third-party judgments to an author’s self-reported private state, authors focused on the discrepancy between the publicly expressed stance—as inferred from social media posts—and the stance measured by public opinion polls. 


\subsection{Emotion Recognition}  

Emotion recognition, equivalently emotion classification, is a widely studied task in NLP that aims to identify the emotions expressed by an author based on their written content \cite{alswaidan2020survey}. Several third-party-annotated datasets have been developed for emotion classification across different domains and applications \cite{liu2019dens, li2017dailydialog, buechel2017readers}. Among these, datasets such as those by \cite{mohammad-etal-2018-semeval} and \cite{demszky-etal-2020-goemotions} are specifically based on social media posts.

Traditional approaches to emotion classification often rely on taxonomies such as Ekman’s six basic emotions \cite{ekman1992argument} or Plutchik’s wheel of emotions \cite{plutchik2001nature}. 
Recently, more nuanced and comprehensive representations have emerged. For example, \cite{demszky-etal-2020-goemotions} introduced a fine-grained emotion taxonomy that includes 27 emotions, along with a neutral category. We adopt this taxonomy in our study. 

Prior psychological studies have shown that the expression and interpretation of emotions can be influenced by cultural and social factors in different ways \cite{lim2016cultural, gendron2014perceptions, mill2009age, barrett2004feelings}. Individuals from the same demographic background—such as age, gender, or cultural identity—tend to exhibit greater alignment in emotion expression and interpretation \cite{elfenbein2002universality, sauter2010cross, elfenbein2002there}. Building on these findings, our study investigates whether human annotators who share demographic traits with an author perform better in emotion recognition tasks compared to those who do not. 

\subsection{Incorporating Social Factors in LLMs} The importance of modeling and incorporating social factors in NLP tasks has been increasingly emphasized.  \cite{hovy-yang-2021-importance} have argued that understanding factors such as the speaker's and receiver's background information will ultimately be necessary if NLP models are to ever achieve human-level performance. Several studies have examined LLMs’ ability to capture and align with the communication styles, perspectives, and preferences of specific demographic groups or even individuals \cite{hwang-etal-2023-aligning, mukherjee-etal-2024-cultural}. 
\cite{beck2024sensitivity} examined socio-demographic prompting, where socio-demographic information is incorporated into prompts to guide an LLM into adopting the perspective of a certain group or user, and found potential in this technique. In contrast, \cite{sun2023aligning} observed that demographic-infused prompts fail to mitigate gender and racial biases in LLMs' annotations. 

\section{Methodology}

To investigate the alignment between first-party labels and third-party annotations in emotion recognition, we adopt the emotion recognition task from \cite{demszky-etal-2020-goemotions} and conduct human subject experiments. Data collection occurred in two stages: (1) collection of first-party social media posts with self-reported emotion labels; and (2) collection of third-party annotations for these posts. 

We analyze annotation performance using Cohen's kappa, F1 score, Recall, and precision. We apply statistical analyses, including mixed linear models and the Wilcoxon signed-rank test, to further assess the impact of demographic similarity on third-party annotation performance. Additionally, we analyze patterns of misalignment to understand the role of linguistic cues in emotion recognition. \textbf{All human subjects research was approved by an Institutional Review Board.}

\subsection{Emotion Taxonomy}
We adopt the fine-grained emotion taxonomy introduced by \citet{demszky-etal-2020-goemotions}, which includes 27 distinct emotions plus ``neutral". The taxonomy provides definitions for each emotion, along with an associated emoji where applicable. The full taxonomy is provided in Appendix \ref{section1}. To evaluate the alignment between third-party annotations and first-party labels at a coarser level, we group the 28 emotion categories into seven groups: joy, love, anger, surprise, fear, sadness, and neutral. As there is no universally agreed-upon set of basic emotions, our grouping is based on the union of basic emotion categories proposed by \citet{ekman1992argument} (anger, disgust, fear, joy, neutral, sadness, surprise) and \citet{shaver1987emotion} (anger, love, fear, joy, sadness, and surprise). Analysis based on this aggregated emotion taxonomy is presented the Appendix \ref{anaysis on aggregated emotion taxonomy}. 

\subsection{First-Party Data Collection}
We recruited social media users through Connect\footnote{https://connect.cloudresearch.com}, a crowdsourcing platform designed to facilitate research participation by connecting researchers with diverse and high-quality participant pools \cite{douglas2023data, eyal2021data}. 
We recruited participants in the United States from three age groups (18–27, 28–43, and 44–59), two gender categories (female and male), and three racial groups (Black, White, and Asian). We attempted to include additional groups, e.g.,  non-binary and Hispanic individuals, but they had insufficient representation on Connect. Our intersectional recruitment strategy aimed to achieve balanced representation across demographics. Specifically, we aimed to obtain approximately 10 responses for each combination of age, gender, and race. While we successfully reached most groups, some intersectional groups—such as Asian participants aged 44–59—proved challenging due to their underrepresentation on the platform. A full breakdown of the number of participants by age, gender, and race is offered in Table \ref{tab:num_participants_posts} of Appendix \ref{section2}. 

\begin{figure*}[t]
  \includegraphics [width=\textwidth]
  {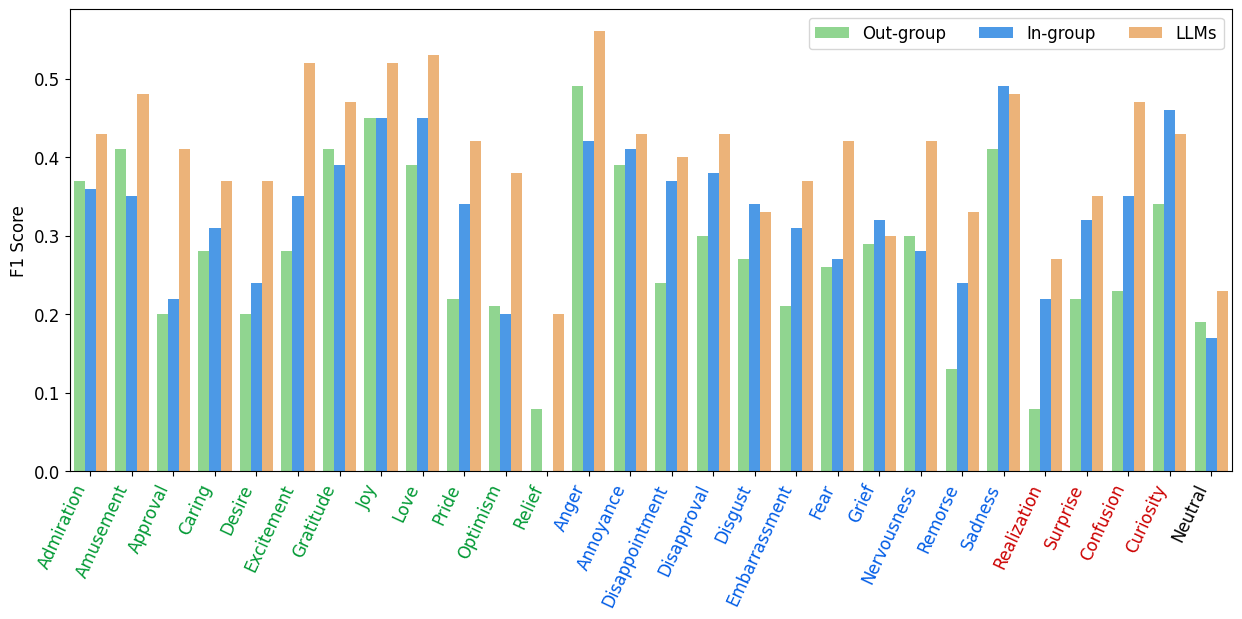}
   \caption{Alignment between third-party annotations with first-party labels, evaluated using F1 score.}
   \label{fig:f1}
 \end{figure*}
 
Via online survey,  
participants were asked to provide basic demographic information and to name one or two social media platforms where they post most frequently. 
They were then asked to upload a minimum of 5 (maximum of 15) posts they had authored on the selected social media platform(s) in the past 12 months. 
Posts were submitted in the form of screenshots. For each uploaded post, participants were asked to review the content and select all emotions they believed were expressed in the post, using the list of 28 emotion categories, along with definitions. To maintain data quality and authenticity, posts were required to adhere to a number of criteria, related to language, multi-media, identifying information, and date of publication. These are detailed in Appendix \ref{criteria}. Each participant was compensated \$2.50 for completing the task, which took approximately 10–15 minutes.

Table~\ref{tab:num_participants_posts} in Appendix \ref{section2} presents the number of participants who provided valid responses and the total number of valid posts for each intersectional group. In total, we collected 729 posts from 123 participants; 44\% of these contained only text and the remaining posts contained both text and images. A quality check was manually performed on these posts and their associated labels; details of this process are provided in Appendix \ref{quality_check}.

\subsection{Third-Party Emotion Labels}
\noindent\textbf{Human annotations.} We recruited human annotators through Connect. For each post, we assigned \textbf{six annotators}: \textbf{three in-group annotators}, who shared all three demographic traits (age group, gender, and racial group) with the author; and, \textbf{three out-group annotators}, who differed from the author on at least two of these traits.
\footnote{As both first-party and third-party human annotators were recruited from the same platform, we ensured that first-party participants did not serve as in-group annotators for their own posts.} 
The demographic distribution of in-group annotators mirrored that of first-party participants by design. 
We aimed to ensure a diverse and balanced sample of out-group annotators across different age groups, genders, and racial backgrounds. Figure \ref{fig:third-party-annotator-demographic} in Appendix \ref{section2} details the demographic distribution of out-group annotators. 

Each annotation task consisted of 5 to 10 posts, which were randomly assigned to annotators. Each annotator was compensated \$2.00 for completing the task, which took approximately 8–10 minutes.  Annotators were presented with the original screenshots of posts provided by their authors and asked to identify the any of emotions expressed by the author of the post (first-party) using the 27 emotion categories plus "neutral" (if none). The list was accompanied by clear definitions of each category. 
In cases where annotators could not confidently assign any labels from the list, they were asked to select "unrecognizable". 
We disregarded annotations that were self-contradictory, such as those which selected both "neutral" and one or more emotions from the list of 27 categories. We interpreted such self-contradictory annotations as indicative of a lack of attention. Furthermore, annotations from annotators who frequently made such errors were removed. Summary statistics of human annotations is presented in Appendix \ref{section2} Table \ref{tab:annotators_statistics}.

\noindent\textbf{LLM annotations.} To complement human annotations, we used LLMs to generate third-party emotion labels. Specifically, we used GPT-4 Turbo \cite{DBLP:journals/corr/abs-2303-08774}, GPT-4o \cite{DBLP:journals/corr/abs-2303-08774}, Gemini 1.5 Pro \cite{DBLP:journals/corr/abs-2403-05530}, Gemini 1.5 Flash \cite{DBLP:journals/corr/abs-2403-05530}, and Claude 3.5 Sonnet \cite{TheC3}. These models were chosen based on their strong performance on other NLP tasks and their ability to process multimodal content. 

Each post was independently annotated by each LLM. LLMs were provided the screenshots of social media posts submitted by study participants, along with instructions defining the 27 emotion categories plus “neutral” and  ``unrecognizable''. The prompt mirrored the instructions given to human annotators to maintain consistency. 

\section{Third-party Annotators' Performance}

\subsection{Third-party Annotators' Agreement}
We estimated interrater agreement amongst human annotators and amongst LLMs, following the method used in \cite{demszky-etal-2020-goemotions}, 
i.e., calculating the Spearman correlation between each annotator's judgments and the mean judgments of other annotators, averaged across all posts labeled by the annotator.  LLMs generally exhibit higher and more consistent interrater agreement compared to human annotators, but there is some variability in this with respect to positive vs. negative emotions. These results are further detailed in Appendix \ref{agreement}.

\subsection{First- and Third-Party Alignment (RQ1)}  
\label{aligment}
We analyze the alignment between first-party labels and third-party annotations. First-party labels are treated as the gold standard. As mentioned earlier, for each post, we collected 6 annotations from human annotators (3 in-group; 3 out-group) and one annotation from each of 5 LLMs. Each annotation provided by a third-party annotator for a post consists of a binary label for each of the 27 emotions plus "neutral". We employed two sets of metrics: (1) Cohen's kappa and (2) classification metrics: F1, recall, and precision.

To assess alignment, we compute Cohen's kappa for both human and LLM annotations relative to the first-party labels. We aggregate annotations for each post using majority voting—applied separately to human annotators and to LLMs. For human annotations, ties can occur after removal of low-quality annotations.
Both human annotators and LLMs exhibit low to fair alignment with first-party labels; their Cohen’s kappa scores range from 0 to 0.45. Details are provided in Appendix \ref{alignment_appendix}.

We further evaluated first- and third-part alignment using F1 score, recall, and precision--standard metrics for multi-label classification tasks. 
The macro-average precision, recall, and F1-score for in-group annotators are (0.38, 0.29, 0.32); for out-group annotators, they are (0.36, 0.24, 0.28). LLMs achieve (0.38, 0.50, 0.40)(see Figure \ref{fig:f1}). 
F1 scores for LLMs range from 0.2 to 0.6, while the F1 scores fall between 0.1 and 0.5 for human. Overall, LLMs outperform human annotators across most emotions. However, for grief, sadness, and curiosity, in-group human annotators demonstrate comparable or even superior performance. Realization, relief, and neutral exhibit consistently low F1 scores across all annotator types, suggesting that these emotions are particularly challenging to classify.  We further examined the correlation between Cohen's kappa and F1 scores for both LLM and human annotations. In both cases, Cohen's kappa scores show a strong positive correlation with F1 scores (LLM: $r = 0.918$, $p \approx 5.86e^{-10}$; 
human: $r = 0.855$, $p \approx 6.85e^{-9}$). These findings highlight that third-party annotations fail to capture first-party intended emotional expressions. We further present confusion matrices showing the alignment between third-party annotations and first-party labels in Appendix \ref{alignment_appendix}, along with detailed analyses.

\subsection{In- vs. Out-Group Annotators (RQ2)}
\label{In-Group VS. Out-group}

To address RQ2, we compare the performance of in- and out-group annotators at both the post and annotator levels. 

\noindent\textbf{Post-level comparison.} Among 91\% posts, both groups reached a majority decision for all 28 emotions. For each post, we computed F1, recall, and precision by comparing the majority-voted labels to the corresponding first-party labels for in-group and out-group. We further adopted Cohen's kappa to assess in-group and out-group's annotation performance. For each emotion, we computed Cohen's kappa based on posts where the majority of annotators in both groups reached a decision on that emotion. On average, 99\% of posts were included for the calculation. 
We perform Wilcoxon signed-rank tests to assess the statistical significance of difference between in-group and out-group performance measured by F1, recall, precision, and recall. Table \ref{tab:ingroup_outgroup_comparison} presents the results of Wilcoxon signed-rank tests. Results indicate that the observed differences in performance of in-group and out-group annotations, measured based on F1, recall, precision, and Cohen's kappa, are statistically significant, allowing us to conclude in-group outperform out-group annotators in recognizing the emotions expressed by first-party authors.
Further details are provided in Figure \ref{fig:f1_dist} and  Figure \ref{fig:cohenKappa_in_out} in Appendix.

\noindent \textbf{Annotator-level comparison} While post-level analysis provides insights into aggregated group performance, annotator-level evaluation helps identify individual alignment with first-party labels. This approach accounts for individual annotator tendencies, which may not be apparent when only considering majority labels. 
To compare in-group and out-group performance at annotator-level, we persevered individual third-party annotations. For each annotation, we obtained F1, recall, and precision by comparing it to the corresponding first-party labels. We averaged the F1 score, precision, and recall across annotations provided within each annotation task for each annotator. 

Figure \ref{fig:f1_dist} in Appendix presents the distribution of average F1 scores, recall, and precision for individual annotators per task, grouped by in-group and out-group membership. Across all three metrics (precision, recall, and F1), the in‐group annotators’ distributions appear shifted slightly to the right compared to the out‐group annotators. The gap is most pronounced in recall and F1, whereas precision exhibits substantial overlap between the two groups. To formally test whether these differences are statistically significant, we employed linear mixed models to account for the nested structure of the data. Specifically, task ID and annotator ID were treated as random effects to account for variability across tasks and individual annotators.

\begin{table}[t]
    \centering
    \begin{tabular}{lccc}
        \hline
        \textbf{Metric} & \makecell{\textbf{In-group} \\ \textbf{Median}} & \makecell{\textbf{Out-group} \\ \textbf{Median}} & \textbf{P-value}  \\
        \hline
        F1 Score      & 0.29  & 0.00  & 0.004*   \\
        Recall        & 0.25  & 0.00  & 0.001* \\
        Precision     & 0.25  & 0.00  & 0.050    \\
        \makecell{Cohen's \\Kappa}  & 0.28  & 0.24  & 0.028* \\
        \hline
    \end{tabular}
    \caption{Comparison of performance between in- and out-group annotators at post-level}
    \label{tab:ingroup_outgroup_comparison}
\end{table}

\begin{table}[t]
\centering

\begin{tabular}{lccl}
\hline
 \textbf{Metric} & \makecell{\textbf{In-group} \\ \textbf{Mean}} & \makecell{\textbf{Out-group} \\ \textbf{Mean}} & \textbf{P-value} \\

\hline
 Precision & 0.32 & 0.30  &  0.052              \\
 F1 Score & 0.30  &  0.28 & 0.023*            \\
Recall  & 0.39  & 0.35  &  0.011*          \\
\hline
\end{tabular}
\caption{Comparison of performance between in- and out-group annotators at annotator-level}
\label{tab:in-out-annotator}
\end{table}

The results of the mixed linear model, presented in Table \ref{tab:in-out-annotator}, indicate that in-group annotators outperform out-group annotators in terms of recall and F1 score. However, the difference in precision between the two groups is not statistically significant. These findings confirm that shared identity traits (age, gender, and race) enhance annotators' ability to align with first-party expressed emotions, particularly in achieving higher recall and overall performance (as measured by F1). 



\subsection{Human Annotators vs. LLMs } 

\begin{table*}[h]
    \centering
    \small
    \begin{tabular}{lccc|ccc}
        \toprule
        \textbf{Metric} & \multicolumn{3}{c|}{\textbf{In-group vs. LLMs}} & \multicolumn{3}{c}{\textbf{Out-group vs. LLMs}} \\
        \cmidrule(lr){2-4} \cmidrule(lr){5-7}
        & \makecell{\textbf{In-group} \\ \textbf{Median}} 
    & \makecell{\textbf{LLM} \\ \textbf{Median}} 
    & \textbf{p-value} 
    & \makecell{\textbf{Out-group} \\ \textbf{Mean}} 
    & \makecell{\textbf{LLM} \\ \textbf{Mean}} 
    & \textbf{p-value} \\
        \midrule
        F1 Score & 0.29 & 0.40 & $8.34 \times 10^{-12}$*** & 0.00 & 0.40 & $4.62 \times 10^{-25}$*** \\
        Recall & 0.25 & 0.33 & $4.95 \times 10^{-31}$*** & 0.00 & 0.33 & $2.76 \times 10^{-39}$*** \\
        Precision & 0.25 & 0.67 & 0.71 & 0.00 & 0.67 & $3.00 \times 10^{-3}$**  \\
        Cohen’s Kappa & 0.28 & 0.32 & $2.75 \times 10^{-5}$*** & 0.23 & 0.32 & $7.45 \times 10^{-9}$*** \\
        \bottomrule
    \end{tabular}
        \caption{Comparison of in- and out-group annotators with LLMs.}
    \label{tab:comparsion}
\end{table*}

We compare the performance of human annotators (both in-group and out-group) with that of LLMs by analyzing F1, recall, and precision metrics on a per-post basis. We compute performance metrics based on the majority-voted labels for each post for in-group, out-group, and LLMs. Comparison between LLMs and in-group is based on posts where the majority of in-group annotators reached a decision for all 28 emotions (94\% of posts). Comparison between LLMs and out-group is based on posts where the majority of out-group annotators reached a decision for all 28 emotions (97\% of posts).

For each emotion, we computed Cohen's kappa for in-group annotations using only posts where a majority of in-group annotators reached a decision (on average, 99\% of posts). Similarly, Cohen's kappa for out-group annotations was calculated using posts where a majority of out-group annotators reached a decision (nearly all posts). Although LLMs provide decisions for all emotions on every post (since we aggregate responses from 5 LLMs), we restricted LLM evaluations to the same subsets of posts used for human annotators to allow direct comparison. 

We employed the Wilcoxon signed-rank test to assess whether differences between in-group and out-group in terms of F1, recall, precision, Cohen's kappa are statistically significant. The results, in Table \ref{tab:comparsion}, suggest that LLMs perform significantly better in terms of F1, recall, and Cohen's kappa. However, their advantage does not extend to precision, as indicated by the lack of statistical significance in the in-group comparison.

\subsection{Demographic Prompting (RQ3)} In this section, we examine the impact of incorporating first-party demographic information—specifically age, gender, and race—into the prompt. By explicitly including demographic details of first-party, we investigate whether the model’s predictions align more closely with first-party emotion labels.

For each LLM, we generated an annotation for every post using demographic prompting. We applied majority voting across the annotations to derive a majority-voted label for each post. We then compared these majority-voted labels with those obtained without demographic information in the prompt. We adopted the Wilcoxon signed-rank test to evaluate whether incorporating demographic information in the prompt improved alignment between the model’s predictions and the first-party labels. 
The results indicate that demographic prompting results in a statistically significant difference in F1 score (p = 0.0095) and precision (p = 0.0004), while no significant difference is observed in recall (p = 0.9934). However, the similar median (F1 score: 0.4, precision: 0.333, recall: 0.667)  and distribution presented in Figure \ref{fig:llm_llm_info} in Appendix, suggesting that while demographic information systematically impacts model predictions, the practical improvement in performance remains minimal.

\section{Understanding Misalignment between First-Party Labels and Third-Party Annotations}
We conducted a qualitative analysis to explore how the linguistic and contextual characteristics of posts may contribute to annotation discrepancies. We examined both cases where third-party annotations closely aligned with first-party labels and cases where they diverged significantly, aiming to identify patterns in post content that may explain these differences. Specifically, we defined high-alignment cases as posts where the majority-voted label received an F1-score greater than 0.6 and low-alignment cases as those with an F1-score below 0.2. These thresholds were determined based on the distribution of F1-scores across annotations. Content within these categories were manually reviewed by two authors.


For posts where third-party annotations (in-group, out-group, and LLMs) achieved near-perfect F1 scores, emotional expression was conveyed through explicit and unambiguous language. For instance, posts categorized by the first party as expressing "joy" explicitly included the word "happy," while expressions of "gratitude" featured phrases such as "thank you," and instances of "confusion" were marked by direct questions. In posts where third-party annotations achieved moderate to high F1 scores, a lack of clear linguistic cues led to subtle discrepancies rather than complete mismatches between the emotions identified by the author and those recognized by third-party annotators. E.g., third-party annotators selected additional emotions that the author did not specify, while in others, certain emotions identified by the author were not acknowledged by annotators.

Among posts exhibiting significant misalignment, we observed three notable patterns among posts: (1) posts that lacked strong textual cues for emotion. For instance, a first-party (author) wrote: "Introducing Toki! :) He's a ringneck dove." and labeled the post with love, excitement, and optimism. (2) A considerable number of posts originated from discussions on Reddit, where the lack of conversational context may have posed a challenge for accurate emotion classification, and (3) many instances involved authors self-reporting "neutral" despite the presence of discernible emotional cues, suggesting that the presence of emotional language does not necessarily reflect the author’s internal emotional state. In addition, we observed that the majority of posts with first-party labels identified as spurious (illustrated in Appendix \ref{quality_check}) were found within this subset of misaligned posts, which is expected.

\section{Conclusion and Discussion}
Our work challenges the assumption that third-party annotations (both human and LLM-based) can reliably infer authors' private states from their texts. By demonstrating a significant misalignment between third-party annotations and first-party (author-provided) labels, we show that third-party annotations fail to accurately capture authors' emotional expressions. We further explored methods to improve third-party annotation quality, leveraging first-party demographic information. We find that demographic similarity between first-party and third-party human annotators enhances annotation performance. Prompting first-party demographic traits marginally enhances LLMs' annotation performance. 

A simple sentence like ``I got a cup of coffee.'' can express different emotions depending on the speaker and the context, information which may or may not be transparent to a third party.  
Similarly, a statement like ``I love morning classes.'' could be interpreted as expressing joy (genuine appreciation) or annoyance (sarcasm or frustration). However, the important question here is: \emph{Are we modeling a third-party's perception of the emotion expressed by the author, or the actual emotion expressed by author in their written text?} 

Third-party annotations struggle to differentiate between semantically similar emotions, such as joy and excitement, and to recognize the complexity of emotional expression—such as frustration underlying gratitude or anger coexisting with caring. This challenge highlights the inherent subjectivity in emotion recognition, where annotators bring their own interpretations, biases, and contextual assumptions to the task. Unlike first-party authors who experience the emotion firsthand, third-party annotators rely solely on textual cues, which may lack sufficient context for accurate inference. As a result, subtle emotional nuances, mixed emotions, or sarcasm often go unnoticed or misinterpreted. Moreover, third-party annotations may reflect individually, socially, culturally influenced perceptions of emotion rather than the actual emotional state intended by the author. For instance, an indirect expression of distress might not be recognized as sadness if it does not conform to expected linguistic patterns. This misalignment could undermine the reliability of models trained on third-party annotations and lead to unintended harm, particularly in high-stakes contexts, e.g., chatbots. 

While our study focuses on emotion recognition, the challenges we identify have implications for other NLP tasks that attempt to infer authors' private states from text. Given these challenges, we argue that it is crucial to critically evaluate the extent to which third-party annotations can serve as a reliable ground truth for modeling authors' private states. Future research should explore methods, such as incorporating first-party first-party feedback and explanation, to achieve  a more truthful representation of the emotions actually expressed by authors. 

\section{Limitations}
The observed limitations of third-party annotations are based on first-party labels and third-party annotations collected from study participants. Even though we implemented strict quality control for first-party posts, we still observed first-party labels that may be spurious. These first-party data were retained in the analysis as whether first-party labels faithfully represent first-party internal emotion can not be externally verifiable. These first-party data were retained in the analysis because whether first-party labels faithfully represent the author's internal emotions cannot be externally verified. In addition, the inclusion of these data points is expected to have minimal impact on the overall findings. 

For third-party annotations, we removed annotations that did not follow annotation instruction, indicating a lack of attention. However,  despite these quality control measures, some low-quality annotations may still remain, as inconsistencies in subjective judgment cannot be entirely eliminated. 

In addition, we lack sufficient data points within each demographic subgroup to conduct statistically robust analyses on how specific identity traits influence annotation performance. Furthermore, while this study examines differences between in-group and out-group annotators, our findings may not generalize to broader populations, cultures, or languages.

\section{Ethical Consideration}

The intention of our work is to highlight critical ethical concerns in inferring authors’ private states (specifically emotions) via third-party annotations. However, there are ethical considerations within the work that need to be considered for further research development. The key issues include privacy risks when handling sensitive emotional data, potential bias from demographic mismatches between authors and annotators, and the unreliability of third-party labels (human or LLM) in capturing genuine emotions, risking harmful misjudgments in real-world applications. While our results show that LLMs outperform humans, in certain scenarios, they may perpetuate training data biases and foster overconfidence in flawed systems. Our study also underscores transparency gaps in methodology and LLM decision-making, labor concerns around displacing human annotators, and risks of mishandling demographic data. It calls for diverse annotation teams, participatory design, rigorous consent protocols, and bias audits to ensure ethical practices in modeling private states. 

\bibliography{custom}

\begin{thebibliography}{62}
\providecommand{\natexlab}[1]{#1}

\bibitem[{AlDayel and Magdy(2021)}]{aldayel2021stance}
Abeer AlDayel and Walid Magdy. 2021.
\newblock Stance detection on social media: State of the art and trends.
\newblock \emph{Information Processing \& Management}, 58(4):102597.

\bibitem[{Alswaidan and Menai(2020)}]{alswaidan2020survey}
Nourah Alswaidan and Mohamed El~Bachir Menai. 2020.
\newblock A survey of state-of-the-art approaches for emotion recognition in text.
\newblock \emph{Knowledge and Information Systems}, 62(8):2937--2987.

\bibitem[{Aman and Szpakowicz(2007)}]{aman2007identifying}
Saima Aman and Stan Szpakowicz. 2007.
\newblock Identifying expressions of emotion in text.
\newblock In \emph{International Conference on Text, Speech and Dialogue}, pages 196--205. Springer.

\bibitem[{Anthropic(2024)}]{TheC3}
Anthropic. 2024.
\newblock The claude 3 model family: Opus, sonnet, haiku.
\newblock Technical report, Anthropic.

\bibitem[{Balahur et~al.(2012)Balahur, Hermida, and Montoyo}]{balahur2012detecting}
Alexandra Balahur, Jes{\'u}s~M Hermida, and Andr{\'e}s Montoyo. 2012.
\newblock Detecting implicit expressions of emotion in text: A comparative analysis.
\newblock \emph{Decision support systems}, 53(4):742--753.

\bibitem[{Bamman and Smith(2015)}]{bamman2015contextualized}
David Bamman and Noah Smith. 2015.
\newblock Contextualized sarcasm detection on twitter.
\newblock In \emph{proceedings of the international AAAI conference on web and social media}, volume~9, pages 574--577.

\bibitem[{Banfield(1982)}]{Banfield1982UnspeakableS}
Ann Banfield. 1982.
\newblock \href {https://api.semanticscholar.org/CorpusID:62165816} {Unspeakable sentences : Narration and representation in the language of fiction}.

\bibitem[{Barrett(2004)}]{barrett2004feelings}
Lisa~Feldman Barrett. 2004.
\newblock Feelings or words? understanding the content in self-report ratings of experienced emotion.
\newblock \emph{Journal of personality and social psychology}, 87(2):266.

\bibitem[{Bauer et~al.(2003)Bauer, Stennes, and Haight}]{bauer2003representation}
Patricia Bauer, Leif Stennes, and Jennifer Haight. 2003.
\newblock Representation of the inner self in autobiography: Women's and men's use of internal states language in personal narratives.
\newblock \emph{Memory}, 11(1):27--42.

\bibitem[{Beck et~al.(2024)Beck, Schuff, Lauscher, and Gurevych}]{beck2024sensitivity}
Tilman Beck, Hendrik Schuff, Anne Lauscher, and Iryna Gurevych. 2024.
\newblock Sensitivity, performance, robustness: Deconstructing the effect of sociodemographic prompting.
\newblock In \emph{Proceedings of the 18th Conference of the European Chapter of the Association for Computational Linguistics (Volume 1: Long Papers)}, pages 2589--2615.

\bibitem[{Brill(1994)}]{brill1994some}
Eric Brill. 1994.
\newblock Some advances in transformation-based part of speech tagging.
\newblock \emph{arXiv preprint cmp-lg/9406010}.

\bibitem[{Brown et~al.(2020)Brown, Mann, Ryder, Subbiah, Kaplan, Dhariwal, Neelakantan, Shyam, Sastry, Askell et~al.}]{brown2020language}
Tom Brown, Benjamin Mann, Nick Ryder, Melanie Subbiah, Jared~D Kaplan, Prafulla Dhariwal, Arvind Neelakantan, Pranav Shyam, Girish Sastry, Amanda Askell, et~al. 2020.
\newblock Language models are few-shot learners.
\newblock \emph{Advances in neural information processing systems}, 33:1877--1901.

\bibitem[{Buechel and Hahn(2017)}]{buechel2017readers}
Sven Buechel and Udo Hahn. 2017.
\newblock Readers vs. writers vs. texts: Coping with different perspectives of text understanding in emotion annotation.
\newblock In \emph{Proceedings of the 11th linguistic annotation workshop}, pages 1--12.

\bibitem[{Chiril et~al.(2022)Chiril, Pamungkas, Benamara, Moriceau, and Patti}]{chiril2022emotionally}
Patricia Chiril, Endang~Wahyu Pamungkas, Farah Benamara, V{\'e}ronique Moriceau, and Viviana Patti. 2022.
\newblock Emotionally informed hate speech detection: a multi-target perspective.
\newblock \emph{Cognitive Computation}, pages 1--31.

\bibitem[{Davidson et~al.(2017)Davidson, Warmsley, Macy, and Weber}]{davidson2017automated}
Thomas Davidson, Dana Warmsley, Michael Macy, and Ingmar Weber. 2017.
\newblock Automated hate speech detection and the problem of offensive language.
\newblock In \emph{Proceedings of the international AAAI conference on web and social media}, volume~11, pages 512--515.

\bibitem[{DeAndrea et~al.(2010)DeAndrea, Shaw, and Levine}]{deandrea2010online}
David~C DeAndrea, Allison~S Shaw, and Timothy~R Levine. 2010.
\newblock Online language: The role of culture in self-expression and self-construal on facebook.
\newblock \emph{Journal of language and social psychology}, 29(4):425--442.

\bibitem[{Demszky et~al.(2020)Demszky, Movshovitz-Attias, Ko, Cowen, Nemade, and Ravi}]{demszky-etal-2020-goemotions}
Dorottya Demszky, Dana Movshovitz-Attias, Jeongwoo Ko, Alan Cowen, Gaurav Nemade, and Sujith Ravi. 2020.
\newblock \href {https://doi.org/10.18653/v1/2020.acl-main.372} {{G}o{E}motions: A dataset of fine-grained emotions}.
\newblock In \emph{Proceedings of the 58th Annual Meeting of the Association for Computational Linguistics}, pages 4040--4054, Online. Association for Computational Linguistics.

\bibitem[{Ding et~al.(2022)Ding, You, Machulla, Jacobs, Sen, and H{\"o}llerer}]{ding2022impact}
Yi~Ding, Jacob You, Tonja-Katrin Machulla, Jennifer Jacobs, Pradeep Sen, and Tobias H{\"o}llerer. 2022.
\newblock Impact of annotator demographics on sentiment dataset labeling.
\newblock \emph{Proceedings of the ACM on Human-Computer Interaction}, 6(CSCW2):1--22.

\bibitem[{Douglas et~al.(2023)Douglas, Ewell, and Brauer}]{douglas2023data}
Benjamin~D Douglas, Patrick~J Ewell, and Markus Brauer. 2023.
\newblock Data quality in online human-subjects research: Comparisons between mturk, prolific, cloudresearch, qualtrics, and sona.
\newblock \emph{Plos one}, 18(3):e0279720.

\bibitem[{Ekman(1992)}]{ekman1992argument}
Paul Ekman. 1992.
\newblock An argument for basic emotions.
\newblock \emph{Cognition \& emotion}, 6(3-4):169--200.

\bibitem[{Elfenbein and Ambady(2002{\natexlab{a}})}]{elfenbein2002there}
Hillary~Anger Elfenbein and Nalini Ambady. 2002{\natexlab{a}}.
\newblock Is there an in-group advantage in emotion recognition?

\bibitem[{Elfenbein and Ambady(2002{\natexlab{b}})}]{elfenbein2002universality}
Hillary~Anger Elfenbein and Nalini Ambady. 2002{\natexlab{b}}.
\newblock On the universality and cultural specificity of emotion recognition: a meta-analysis.
\newblock \emph{Psychological bulletin}, 128(2):203.

\bibitem[{Eyal et~al.(2021)Eyal, David, Andrew, Zak, and Ekaterina}]{eyal2021data}
Peer Eyal, Rothschild David, Gordon Andrew, Evernden Zak, and Damer Ekaterina. 2021.
\newblock Data quality of platforms and panels for online behavioral research.
\newblock \emph{Behavior research methods}, pages 1--20.

\bibitem[{Gendron et~al.(2014)Gendron, Roberson, van~der Vyver, and Barrett}]{gendron2014perceptions}
Maria Gendron, Debi Roberson, Jacoba~Marietta van~der Vyver, and Lisa~Feldman Barrett. 2014.
\newblock Perceptions of emotion from facial expressions are not culturally universal: evidence from a remote culture.
\newblock \emph{Emotion}, 14(2):251.

\bibitem[{Gilardi et~al.(2023)Gilardi, Alizadeh, and Kubli}]{DBLP:journals/corr/abs-2303-15056}
Fabrizio Gilardi, Meysam Alizadeh, and Ma{\"{e}}l Kubli. 2023.
\newblock Chatgpt outperforms crowd-workers for text-annotation tasks.
\newblock \emph{CoRR}, abs/2303.15056.

\bibitem[{Hasanain et~al.(2024)Hasanain, Ahmad, and Alam}]{DBLP:conf/emnlp/HasanainAA24}
Maram Hasanain, Fatema Ahmad, and Firoj Alam. 2024.
\newblock Large language models for propaganda span annotation.
\newblock In \emph{{EMNLP} (Findings)}, pages 14522--14532. Association for Computational Linguistics.

\bibitem[{Hovy and Yang(2021)}]{hovy-yang-2021-importance}
Dirk Hovy and Diyi Yang. 2021.
\newblock \href {https://doi.org/10.18653/v1/2021.naacl-main.49} {The importance of modeling social factors of language: Theory and practice}.
\newblock In \emph{Proceedings of the 2021 Conference of the North American Chapter of the Association for Computational Linguistics: Human Language Technologies}, pages 588--602, Online. Association for Computational Linguistics.

\bibitem[{Hwang et~al.(2023)Hwang, Majumder, and Tandon}]{hwang-etal-2023-aligning}
EunJeong Hwang, Bodhisattwa Majumder, and Niket Tandon. 2023.
\newblock \href {https://doi.org/10.18653/v1/2023.findings-emnlp.393} {Aligning language models to user opinions}.
\newblock In \emph{Findings of the Association for Computational Linguistics: EMNLP 2023}, pages 5906--5919, Singapore. Association for Computational Linguistics.

\bibitem[{Iyyer et~al.(2014)Iyyer, Enns, Boyd-Graber, and Resnik}]{iyyer2014political}
Mohit Iyyer, Peter Enns, Jordan Boyd-Graber, and Philip Resnik. 2014.
\newblock Political ideology detection using recursive neural networks.
\newblock In \emph{Proceedings of the 52nd annual meeting of the Association for Computational Linguistics (volume 1: long papers)}, pages 1113--1122.

\bibitem[{Joseph et~al.(2021)Joseph, Shugars, Gallagher, Green, Quintana~Math{\'e}, An, and Lazer}]{joseph-etal-2021-mis}
Kenneth Joseph, Sarah Shugars, Ryan Gallagher, Jon Green, Alexi Quintana~Math{\'e}, Zijian An, and David Lazer. 2021.
\newblock \href {https://doi.org/10.18653/v1/2021.emnlp-main.27} {(mis)alignment between stance expressed in social media data and public opinion surveys}.
\newblock In \emph{Proceedings of the 2021 Conference on Empirical Methods in Natural Language Processing}, pages 312--324, Online and Punta Cana, Dominican Republic. Association for Computational Linguistics.

\bibitem[{Kim et~al.(2024)Kim, Mitra, Li~Chen, Rahman, and Zhang}]{kim-etal-2024-meganno}
Hannah Kim, Kushan Mitra, Rafael Li~Chen, Sajjadur Rahman, and Dan Zhang. 2024.
\newblock \href {https://aclanthology.org/2024.eacl-demo.18/} {{MEGA}nno+: A human-{LLM} collaborative annotation system}.
\newblock In \emph{Proceedings of the 18th Conference of the European Chapter of the Association for Computational Linguistics: System Demonstrations}, pages 168--176, St. Julians, Malta. Association for Computational Linguistics.

\bibitem[{Li and Conrad(2024)}]{li2024advancing}
Mao Li and Frederick Conrad. 2024.
\newblock Advancing annotation of stance in social media posts: A comparative analysis of large language models and crowd sourcing.
\newblock \emph{arXiv preprint arXiv:2406.07483}.

\bibitem[{Li et~al.(2023)Li, Shi, Ziems, Kan, Chen, Liu, and Yang}]{li2023coannotating}
Minzhi Li, Taiwei Shi, Caleb Ziems, Min-Yen Kan, Nancy~F Chen, Zhengyuan Liu, and Diyi Yang. 2023.
\newblock Coannotating: Uncertainty-guided work allocation between human and large language models for data annotation.
\newblock \emph{arXiv preprint arXiv:2310.15638}.

\bibitem[{Li et~al.(2017)Li, Su, Shen, Li, Cao, and Niu}]{li2017dailydialog}
Yanran Li, Hui Su, Xiaoyu Shen, Wenjie Li, Ziqiang Cao, and Shuzi Niu. 2017.
\newblock Dailydialog: A manually labelled multi-turn dialogue dataset.
\newblock \emph{arXiv preprint arXiv:1710.03957}.

\bibitem[{Lim(2016)}]{lim2016cultural}
Nangyeon Lim. 2016.
\newblock Cultural differences in emotion: differences in emotional arousal level between the east and the west.
\newblock \emph{Integrative medicine research}, 5(2):105--109.

\bibitem[{Liu et~al.(2019)Liu, Osama, and De~Andrade}]{liu2019dens}
Chen Liu, Muhammad Osama, and Anderson De~Andrade. 2019.
\newblock Dens: A dataset for multi-class emotion analysis.
\newblock In \emph{Proceedings of the 2019 Conference on Empirical Methods in Natural Language Processing and the 9th International Joint Conference on Natural Language Processing (EMNLP-IJCNLP)}, pages 6293--6298.

\bibitem[{Mill et~al.(2009)Mill, Allik, Realo, and Valk}]{mill2009age}
Aire Mill, J{\"u}ri Allik, Anu Realo, and Raivo Valk. 2009.
\newblock Age-related differences in emotion recognition ability: a cross-sectional study.
\newblock \emph{Emotion}, 9(5):619.

\bibitem[{Mohammad and Bravo-Marquez(2017)}]{mohammad-bravo-marquez-2017-emotion}
Saif Mohammad and Felipe Bravo-Marquez. 2017.
\newblock \href {https://doi.org/10.18653/v1/S17-1007} {Emotion intensities in tweets}.
\newblock In \emph{Proceedings of the 6th Joint Conference on Lexical and Computational Semantics (*{SEM} 2017)}, pages 65--77, Vancouver, Canada. Association for Computational Linguistics.

\bibitem[{Mohammad et~al.(2018)Mohammad, Bravo-Marquez, Salameh, and Kiritchenko}]{mohammad-etal-2018-semeval}
Saif Mohammad, Felipe Bravo-Marquez, Mohammad Salameh, and Svetlana Kiritchenko. 2018.
\newblock \href {https://doi.org/10.18653/v1/S18-1001} {{S}em{E}val-2018 task 1: Affect in tweets}.
\newblock In \emph{Proceedings of the 12th International Workshop on Semantic Evaluation}, pages 1--17, New Orleans, Louisiana. Association for Computational Linguistics.

\bibitem[{Mukherjee et~al.(2024)Mukherjee, Adilazuarda, Sitaram, Bali, Aji, and Choudhury}]{mukherjee-etal-2024-cultural}
Sagnik Mukherjee, Muhammad~Farid Adilazuarda, Sunayana Sitaram, Kalika Bali, Alham~Fikri Aji, and Monojit Choudhury. 2024.
\newblock \href {https://doi.org/10.18653/v1/2024.emnlp-main.884} {Cultural conditioning or placebo? on the effectiveness of socio-demographic prompting}.
\newblock In \emph{Proceedings of the 2024 Conference on Empirical Methods in Natural Language Processing}, pages 15811--15837, Miami, Florida, USA. Association for Computational Linguistics.

\bibitem[{Nadeau and Sekine(2009)}]{nadeau2009survey}
David Nadeau and Satoshi Sekine. 2009.
\newblock A survey of named entity recognition and classification.
\newblock In \emph{Named Entities: Recognition, classification and use}, pages 3--28. John Benjamins publishing company.

\bibitem[{Nemes and Kiss(2021)}]{nemes2021social}
L{\'a}szl{\'o} Nemes and Attila Kiss. 2021.
\newblock Social media sentiment analysis based on covid-19.
\newblock \emph{Journal of Information and Telecommunication}, 5(1):1--15.

\bibitem[{Niu et~al.(2024)Niu, Jaiswal, and Provost}]{DBLP:journals/corr/abs-2408-17026}
Minxue Niu, Mimansa Jaiswal, and Emily~Mower Provost. 2024.
\newblock From text to emotion: Unveiling the emotion annotation capabilities of llms.
\newblock \emph{CoRR}, abs/2408.17026.

\bibitem[{OpenAI(2023)}]{DBLP:journals/corr/abs-2303-08774}
OpenAI. 2023.
\newblock {GPT-4} technical report.
\newblock \emph{CoRR}, abs/2303.08774.

\bibitem[{Oprea and Magdy(2019)}]{oprea2019isarcasm}
Silviu Oprea and Walid Magdy. 2019.
\newblock isarcasm: A dataset of intended sarcasm.
\newblock \emph{arXiv preprint arXiv:1911.03123}.

\bibitem[{Pavlopoulos et~al.(2020)Pavlopoulos, Sorensen, Dixon, Thain, and Androutsopoulos}]{DBLP:conf/acl/PavlopoulosSDTA20}
John Pavlopoulos, Jeffrey Sorensen, Lucas Dixon, Nithum Thain, and Ion Androutsopoulos. 2020.
\newblock Toxicity detection: Does context really matter?
\newblock In \emph{{ACL}}, pages 4296--4305. Association for Computational Linguistics.

\bibitem[{Plutchik(2001)}]{plutchik2001nature}
Robert Plutchik. 2001.
\newblock The nature of emotions: Human emotions have deep evolutionary roots, a fact that may explain their complexity and provide tools for clinical practice.
\newblock \emph{American scientist}, 89(4):344--350.

\bibitem[{Quirk et~al.(1985)Quirk, Greenbaum, Leech, and Svartvik}]{quirk1985grammar}
Randolph Quirk, Sidney Greenbaum, Geoffrey Leech, and Jan Svartvik. 1985.
\newblock \emph{A Comprehensive Grammar of the English Language}.
\newblock Longman, New York.

\bibitem[{Rashtchian et~al.(2010)Rashtchian, Young, Hodosh, and Hockenmaier}]{rashtchian2010collecting}
Cyrus Rashtchian, Peter Young, Micah Hodosh, and Julia Hockenmaier. 2010.
\newblock Collecting image annotations using amazon’s mechanical turk.
\newblock In \emph{Proceedings of the NAACL HLT 2010 workshop on creating speech and language data with Amazon’s Mechanical Turk}, pages 139--147.

\bibitem[{Reid et~al.(2024)Reid, Savinov, Teplyashin, Lepikhin, Lillicrap, Alayrac, Soricut, Lazaridou, Firat, Schrittwieser, Antonoglou, Anil, Borgeaud, Dai, Millican, Dyer, Glaese, Sottiaux, Lee, Viola, Reynolds, Xu, Molloy, Chen, Isard, Barham, Hennigan, McIlroy, Johnson, Schalkwyk, Collins, Rutherford, Moreira, Ayoub, Goel, Meyer, Thornton, Yang, Michalewski, Abbas, Schucher, Anand, Ives, Keeling, Lenc, Haykal, Shakeri, Shyam, Chowdhery, Ring, Spencer, Sezener, and et~al.}]{DBLP:journals/corr/abs-2403-05530}
Machel Reid, Nikolay Savinov, Denis Teplyashin, Dmitry Lepikhin, Timothy~P. Lillicrap, Jean{-}Baptiste Alayrac, Radu Soricut, Angeliki Lazaridou, Orhan Firat, Julian Schrittwieser, Ioannis Antonoglou, Rohan Anil, Sebastian Borgeaud, Andrew~M. Dai, Katie Millican, Ethan Dyer, Mia Glaese, Thibault Sottiaux, Benjamin Lee, Fabio Viola, Malcolm Reynolds, Yuanzhong Xu, James Molloy, Jilin Chen, Michael Isard, Paul Barham, Tom Hennigan, Ross McIlroy, Melvin Johnson, Johan Schalkwyk, Eli Collins, Eliza Rutherford, Erica Moreira, Kareem Ayoub, Megha Goel, Clemens Meyer, Gregory Thornton, Zhen Yang, Henryk Michalewski, Zaheer Abbas, Nathan Schucher, Ankesh Anand, Richard Ives, James Keeling, Karel Lenc, Salem Haykal, Siamak Shakeri, Pranav Shyam, Aakanksha Chowdhery, Roman Ring, Stephen Spencer, Eren Sezener, and et~al. 2024.
\newblock Gemini 1.5: Unlocking multimodal understanding across millions of tokens of context.
\newblock \emph{CoRR}, abs/2403.05530.

\bibitem[{Sauter et~al.(2010)Sauter, Eisner, Ekman, and Scott}]{sauter2010cross}
Disa~A Sauter, Frank Eisner, Paul Ekman, and Sophie~K Scott. 2010.
\newblock Cross-cultural recognition of basic emotions through nonverbal emotional vocalizations.
\newblock \emph{Proceedings of the National Academy of Sciences}, 107(6):2408--2412.

\bibitem[{Shaver et~al.(1987)Shaver, Schwartz, Kirson, and O'connor}]{shaver1987emotion}
Phillip Shaver, Judith Schwartz, Donald Kirson, and Cary O'connor. 1987.
\newblock Emotion knowledge: further exploration of a prototype approach.
\newblock \emph{Journal of personality and social psychology}, 52(6):1061.

\bibitem[{Shaw and Lyons(2017)}]{shaw2017lie}
Hannah Shaw and Minna Lyons. 2017.
\newblock Lie detection accuracy—the role of age and the use of emotions as a reliable cue.
\newblock \emph{Journal of Police and Criminal Psychology}, 32:300--304.

\bibitem[{Shen and Rose(2021)}]{shen-rose-2021-sounds}
Qinlan Shen and Carolyn Rose. 2021.
\newblock \href {https://doi.org/10.18653/v1/2021.eacl-main.152} {What sounds {\textquotedblleft}right{\textquotedblright} to me? experiential factors in the perception of political ideology}.
\newblock In \emph{Proceedings of the 16th Conference of the European Chapter of the Association for Computational Linguistics: Main Volume}, pages 1762--1771, Online. Association for Computational Linguistics.

\bibitem[{Snow et~al.(2008)Snow, O’connor, Jurafsky, and Ng}]{snow2008cheap}
Rion Snow, Brendan O’connor, Dan Jurafsky, and Andrew~Y Ng. 2008.
\newblock Cheap and fast--but is it good? evaluating non-expert annotations for natural language tasks.
\newblock In \emph{Proceedings of the 2008 conference on empirical methods in natural language processing}, pages 254--263.

\bibitem[{Sun et~al.(2023)Sun, Pei, Choi, and Jurgens}]{sun2023aligning}
Huaman Sun, Jiaxin Pei, Minje Choi, and David Jurgens. 2023.
\newblock Aligning with whom? large language models have gender and racial biases in subjective nlp tasks.
\newblock \emph{arXiv preprint arXiv:2311.09730}.

\bibitem[{Venkit et~al.(2023)Venkit, Srinath, Gautam, Venkatraman, Gupta, Passonneau, and Wilson}]{venkit-etal-2023-sentiment}
Pranav Venkit, Mukund Srinath, Sanjana Gautam, Saranya Venkatraman, Vipul Gupta, Rebecca Passonneau, and Shomir Wilson. 2023.
\newblock \href {https://doi.org/10.18653/v1/2023.emnlp-main.848} {The sentiment problem: A critical survey towards deconstructing sentiment analysis}.
\newblock In \emph{Proceedings of the 2023 Conference on Empirical Methods in Natural Language Processing}, pages 13743--13763, Singapore. Association for Computational Linguistics.

\bibitem[{Waterloo et~al.(2018)Waterloo, Baumgartner, Peter, and Valkenburg}]{waterloo2018norms}
Sophie~F Waterloo, Susanne~E Baumgartner, Jochen Peter, and Patti~M Valkenburg. 2018.
\newblock Norms of online expressions of emotion: Comparing facebook, twitter, instagram, and whatsapp.
\newblock \emph{New media \& society}, 20(5):1813--1831.

\bibitem[{Wiebe et~al.(2004)Wiebe, Wilson, Bruce, Bell, and Martin}]{wiebe2004learning}
Janyce Wiebe, Theresa Wilson, Rebecca Bruce, Matthew Bell, and Melanie Martin. 2004.
\newblock Learning subjective language.
\newblock \emph{Computational linguistics}, 30(3):277--308.

\bibitem[{Wiebe et~al.(2005)Wiebe, Wilson, and Cardie}]{wiebe2005annotating}
Janyce Wiebe, Theresa Wilson, and Claire Cardie. 2005.
\newblock Annotating expressions of opinions and emotions in language.
\newblock \emph{Language resources and evaluation}, 39:165--210.

\bibitem[{Zhang et~al.(2023)Zhang, Yang, Ji, and Ananiadou}]{zhang2023emotion}
Tianlin Zhang, Kailai Yang, Shaoxiong Ji, and Sophia Ananiadou. 2023.
\newblock Emotion fusion for mental illness detection from social media: A survey.
\newblock \emph{Information Fusion}, 92:231--246.

\bibitem[{Zygad{\l}o et~al.(2021)Zygad{\l}o, Koz{\l}owski, and Janicki}]{zygadlo2021text}
Artur Zygad{\l}o, Marek Koz{\l}owski, and Artur Janicki. 2021.
\newblock Text-based emotion recognition in english and polish for therapeutic chatbot.
\newblock \emph{Applied Sciences}, 11(21):10146.

\end{thebibliography}

\appendix

\clearpage

\section{Emotion Taxonomy}

\begin{figure}[t]
  \includegraphics[width=\columnwidth]{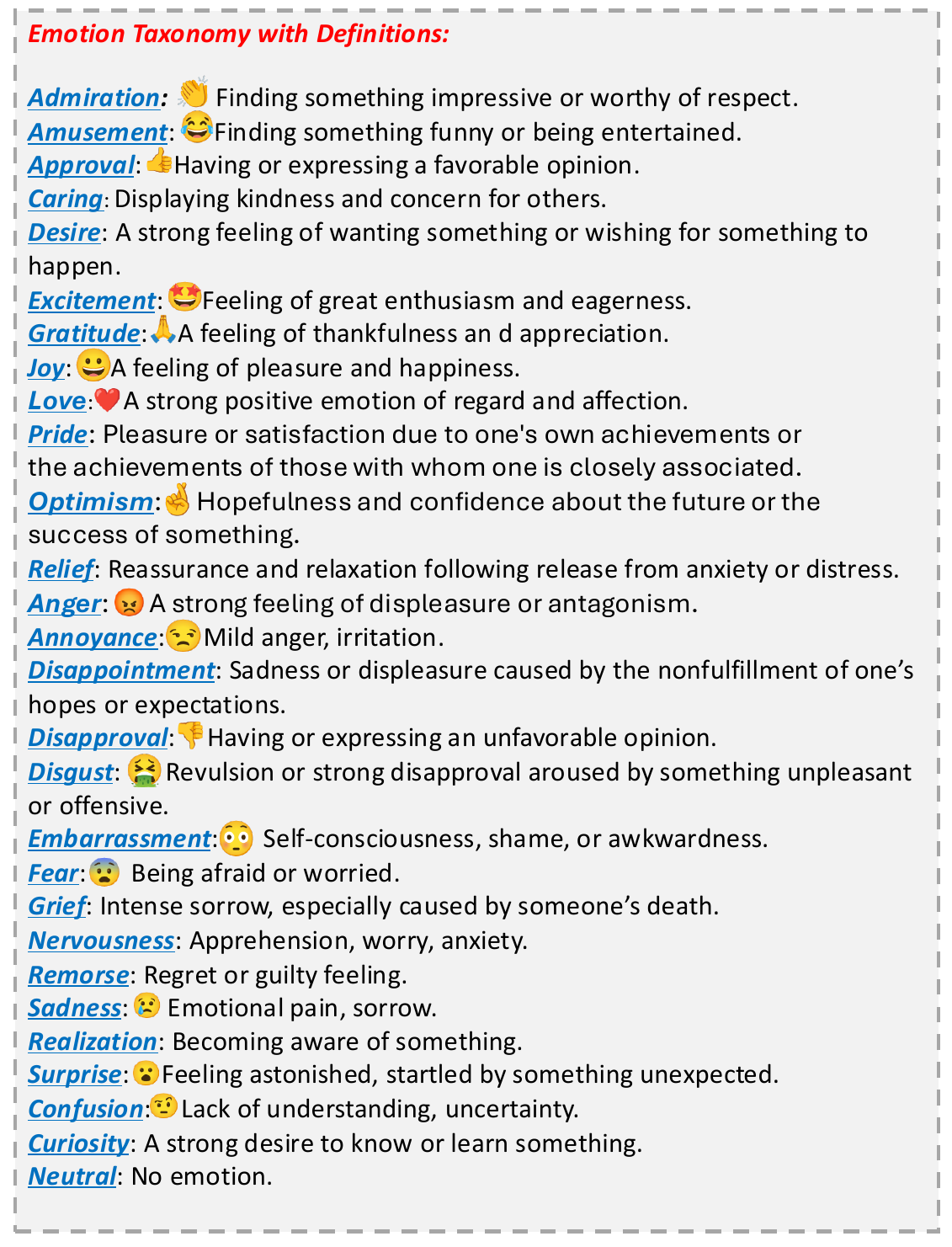}
   \caption{Emotion taxonomy}
   \label{fig:taxonomy}
 \end{figure}

We adopted the emotion taxonomy introduced in \cite{demszky-etal-2020-goemotions}, presented in Figure \ref{fig:taxonomy}.
\label{section1}

\section{Demographics and Annotation Statistics}
\label{section2}
\begin{figure}[h]
  \includegraphics[width=\columnwidth]{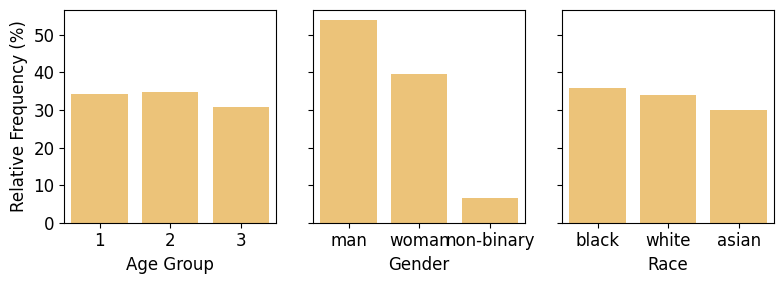}
   \caption{Distribution of third-party annotators by age group, gender, and race.}
   \label{fig:third-party-annotator-demographic}
 \end{figure}

In this section, we present the demographic breakdown of first-party participants in Table \ref{tab:num_participants_posts} and the demographic breakdown of third-party annotators in Figure \ref{fig:third-party-annotator-demographic}. Additionally, Table \ref{tab:annotators_statistics} provides summary statistics of third-party annotations.

\begin{table}[h]
  \centering
  \begin{tabular}{cccccc}
    \hline
    \textbf{Age} & \textbf{Gender} & \textbf{Race} & \textbf{Participants} &  \textbf{Posts} \\
    \hline
    18-27 & Man   & Asian & 8  & 33 \\
    18-27 & Man   & Black & 7  & 55 \\
    18-27 & Man   & White & 13 & 66 \\
    18-27 & Woman & Asian & 10 & 66 \\
    18-27 & Woman & Black & 8  & 48 \\
    18-27 & Woman & White & 8  & 57 \\
    28-43 & Man   & Asian & 8  & 39 \\
    28-43 & Man   & Black & 9  & 48 \\
    28-43 & Man   & White & 8  & 59 \\
    28-43 & Woman & Asian & 9  & 44 \\
    28-43 & Woman & Black & 9  & 51 \\
    28-43 & Woman & White & 9  & 58 \\
    44-59 & Man   & White & 8  & 51 \\
    44-59 & Woman & White & 9  & 54 \\
    \hline
  \end{tabular}
  \caption{Demographic breakdown of first-party participants, showing the number of individuals and posts for each combination of age, gender, and race.}\label{tab:num_participants_posts}
\end{table}

\begin{table}[h]
\begin{tabular}{|p{6cm}|l|}
\hline
\textbf{Number of annotators (unique)} & 399  \\ \hline
\textbf{Number of annotators} & \\ 
\quad - In-group role & 236 \\ 
\quad - Out-group role & 201 \\ \hline
\textbf{Average number of posts per annotator} & \\ 
\quad - In-group role & 9.1 \\ 
\quad - Out-group role & 10.7 \\ \hline
\textbf{Total annotations} & \\ 
\quad - By in-group annotators & 2136 \\ 
\quad - By out-group annotators & 2157 \\ \hline
\end{tabular}
\caption{Summary statistics of third-party annotations.}
\label{tab:annotators_statistics}
\end{table}

\section{Criteria and Verification for First-party Posts}
\label{criteria}

To maintain data quality and genuine responses, submitted posts were required to adhere to the follow criteria:(1) originally created by the participant within the past 12 months (excluding shares, reposts, or non-original content); (2) in English and contain at least 5 words, excluding hashtags and URLs; (3) if multimedia is included, it should contain only images, with emotion conveyed through the images and text captured in the screenshot(s); (4) published at least 24 hours prior to submission; (5) fully captured in the screenshots.
Additionally, participants were required to submit a screenshot of their social media account page, displaying their profile name, with the option to redact other information. The UI features of the account page helped us verify that participants had provided their own social media posts. 
We reviewed all submissions and removed posts that failed to meet one or more of the above criteria. We also excluded the entire response of participants whose submission authenticity could not be verified. 

\begin{figure*}[t]
  \includegraphics[width=\textwidth]{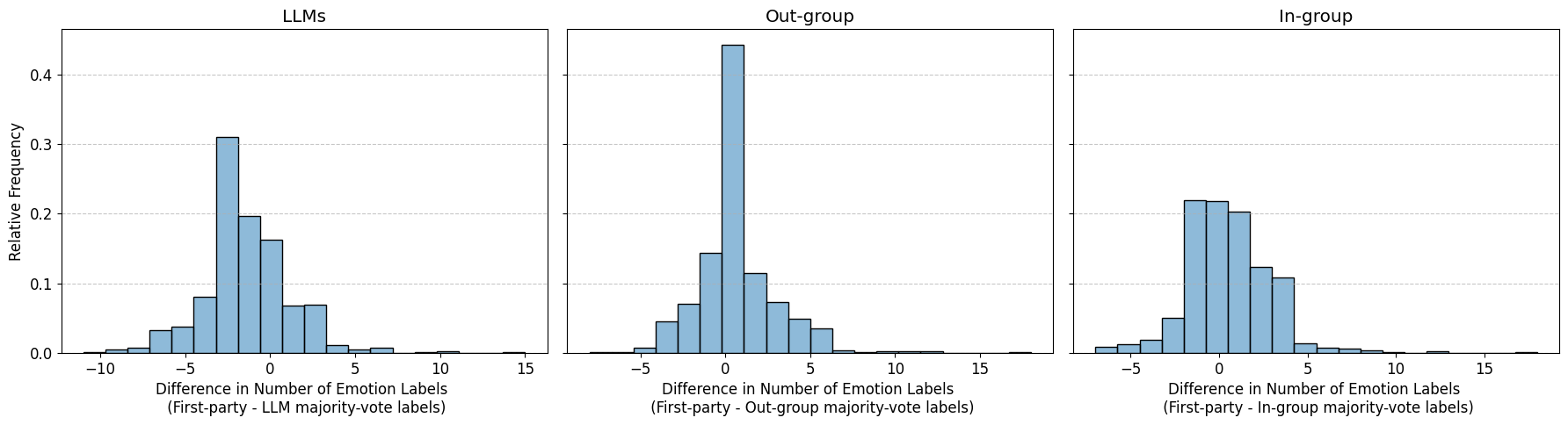}
   \caption{Distributions of the differences in the number of emotion labels assigned by first-party annotators and third-party annotators. }
   \label{fig:diff_num_label_dist}
 \end{figure*}

\begin{figure}[t]
\includegraphics[width=\columnwidth]{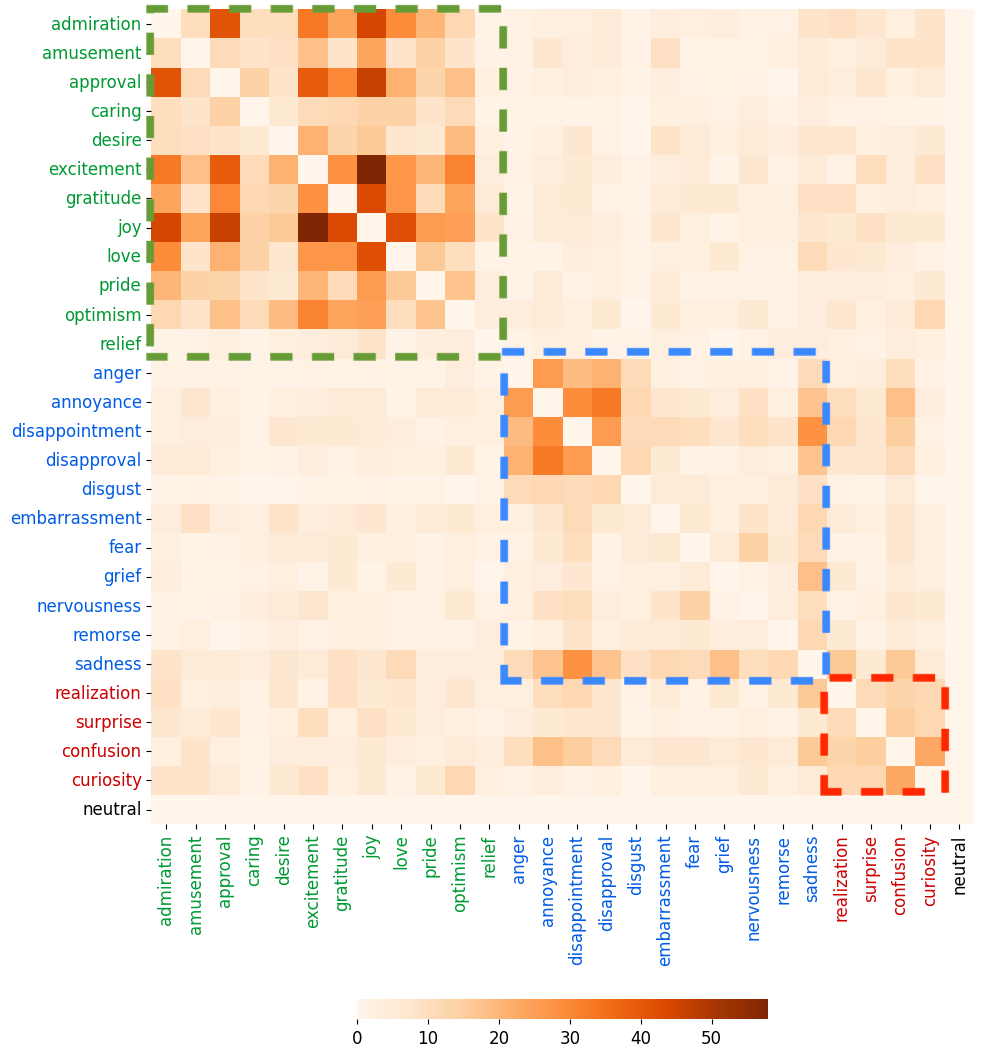}
\caption{Co-occurrence of first-party emotion labels.}
\label{fig:heatmap}
 \end{figure}

\section{Quality Check on First-party Labels}
\label{quality_check}
Two authors separately reviewed each post and its corresponding first-party label to identify spurious labels—cases where the first-party label could not be reasonably justified by the post content. They labeled only clear mismatches between first-party labels and the emotional expression in the posts, rather than cases of subtler misalignment where there might be room for interpretation. They agreed on 89\% of the posts. For posts where disagreements occurred, a senior author conducted a final review to resolve discrepancies.

Among the 729 posts, 9.2\% of first-party labels were flagged as spurious. , suggesting that the majority of first-party labels are reasonably aligned with the post content, indicating high quality. We retain posts identified as spurious in our analysis as a label flagged by us as spurious does not necessarily mean that it fails to accurately reflect the author's internal emotional state. We believe these labels are still valuable, as they highlight the complexity of emotional expression and the challenges of inferring emotions solely from textual content.

\section{Statistical Overview of First-party and Third-party Labels}
\label{stats_overview}
\textbf{First-party labels} 
Among all posts, 6.32\% are labeled as Neutral, indicating no emotion is expressed. Of the remaining posts, 37.9\% have a single emotion label, 25.3\% have two labels, 14.6\% have three labels, and 22.3\% have four or more labels. Figure~\ref{fig:heatmap} shows the co-occurrence patterns of emotion labels, where the labels on the x- and y-axes are color-coded to represent distinct semantic categories: positive, negative, and ambiguous emotions. Positive emotions, such as joy, excitement, and admiration, occur more frequently overall compared to negative emotions like anger or disapproval. The heatmap reveals that emotions with similar conceptual or semantic tones often co-occur within the same post. For instance, positive emotions like joy, excitement, and admiration frequently appear together. Similarly, subsets of negative emotions, such as disappointment and annoyance, also exhibit notable co-occurrence. This pattern aligns with the observed tendency of social media users to predominantly share positive sentiments, while negative emotions appear less frequently \cite{waterloo2018norms}. The co-occurrence of semantically related emotions suggests that users naturally cluster similar emotional tones in their posts, reflecting common patterns in emotional expression.

\begin{figure*}[t]
\includegraphics [width=\textwidth]
   {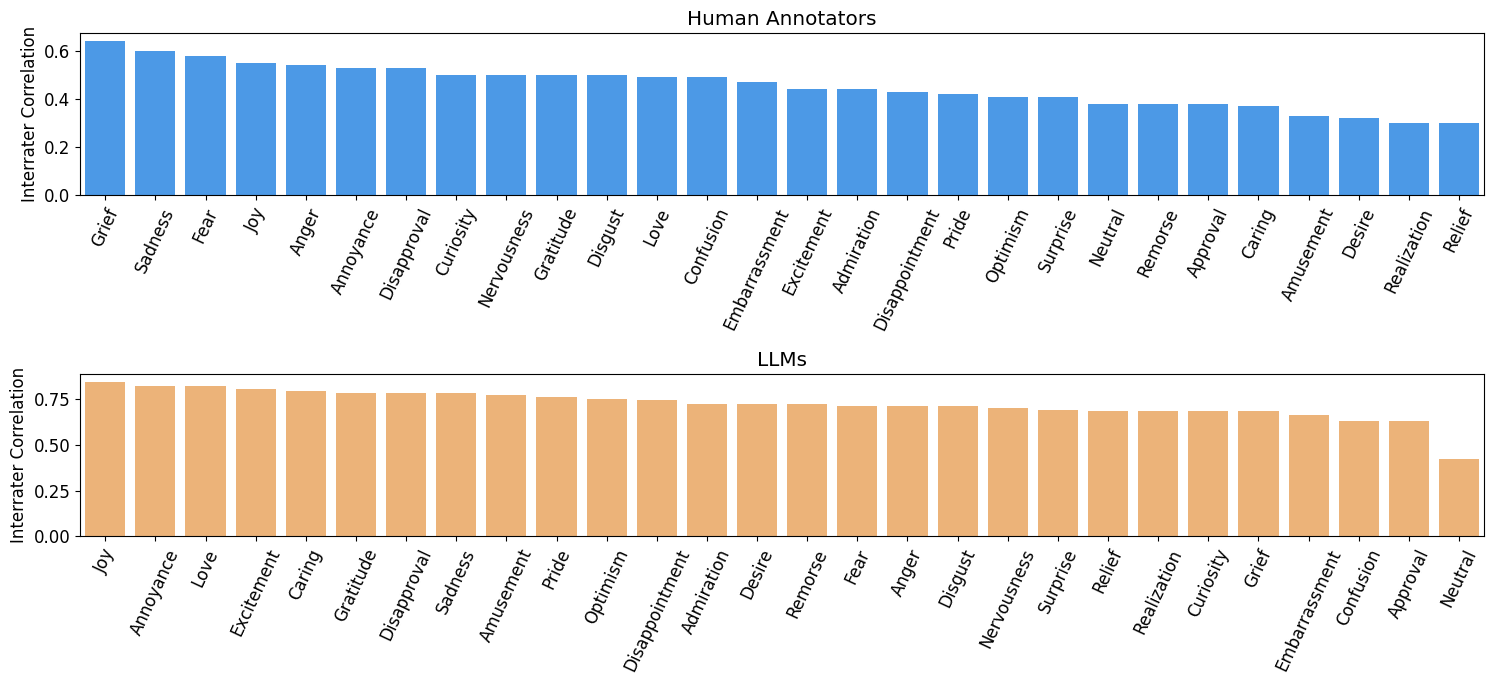}
    \caption{Interrater correlation among third-party annotators.}
    \label{fig:inter_rater}
\end{figure*}

\noindent \textbf{Third-party annotations} We applied majority voting separately to in-group annotations, out-group annotations, and LLM annotations. For in-group annotators, a majority decision was reached for all 28 emotions in 94\% of posts, while for out-group annotators, a majority decision was reached for all 28 emotions in 97\% of posts. In the case of LLMs, a majority label was assigned for all 28 emotions in every post, as we obtained annotations from 5 LLMs. 4.1\% of posts were labeled as neutral by in-group annotators, 4.4\% by out-group annotators, and 0.1\% by LLMs. For each annotator group, for posts not labeled as neutral, we calculated the difference in the number of emotion labels selected by first-party participants (authors) and by third-party annotators. 

Figure \ref{fig:diff_num_label_dist} shows the distribution. The x-axis represents the difference in the number of emotion labels (calculated as first-party minus third-party), where negative values indicate over-labeling by third-party annotators, and positive values indicate under-labeling. LLMs exhibit a wider distribution with more over-labeling, while out-group annotators tend to align more closely with first-party labels, clustering around zero. In-group annotators show slightly more variation but tend to assign fewer labels than first-party participants. 

\begin{figure*}[h]
\includegraphics [width=\textwidth]
{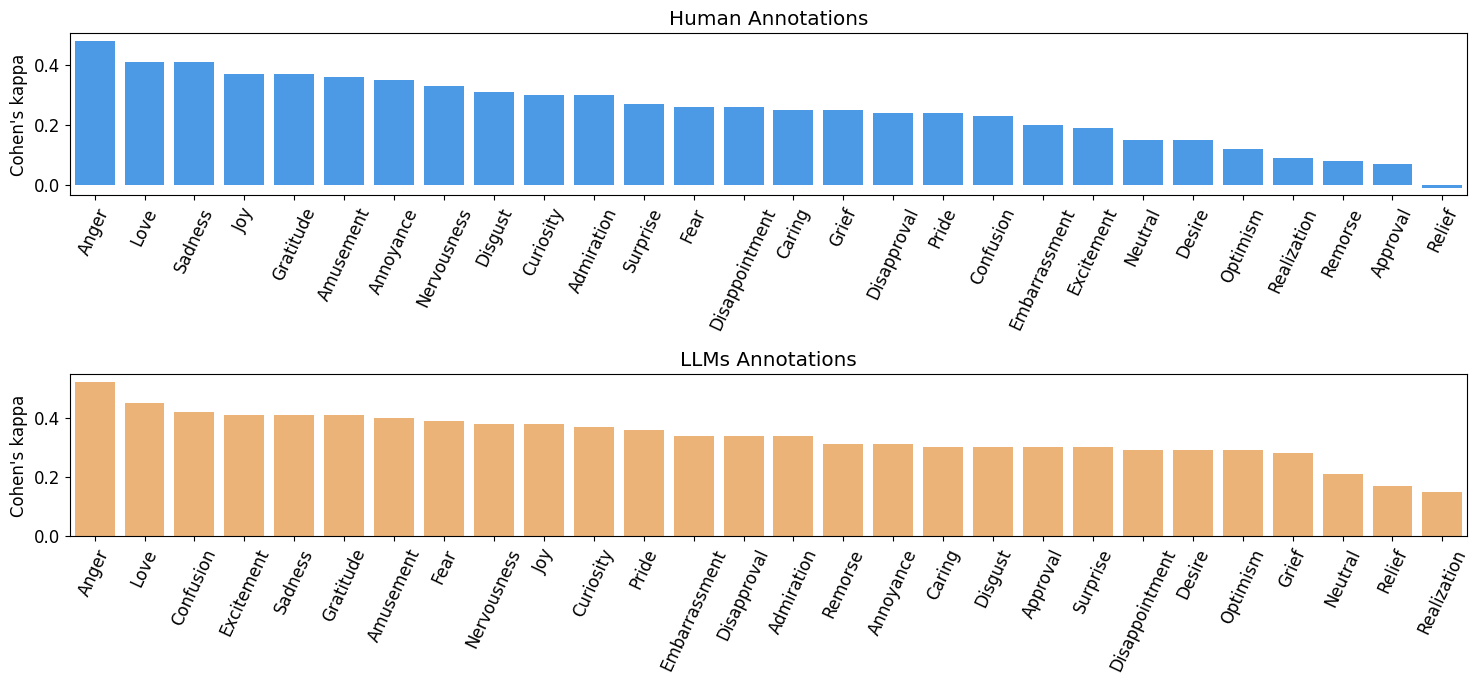}
\caption{Alignment between third-party annotations with first-party labels, evaluated using Cohen's kappa.}
\label{fig:cohenKappa}
\end{figure*}

 \begin{figure}[t]
  \includegraphics[width=\columnwidth]{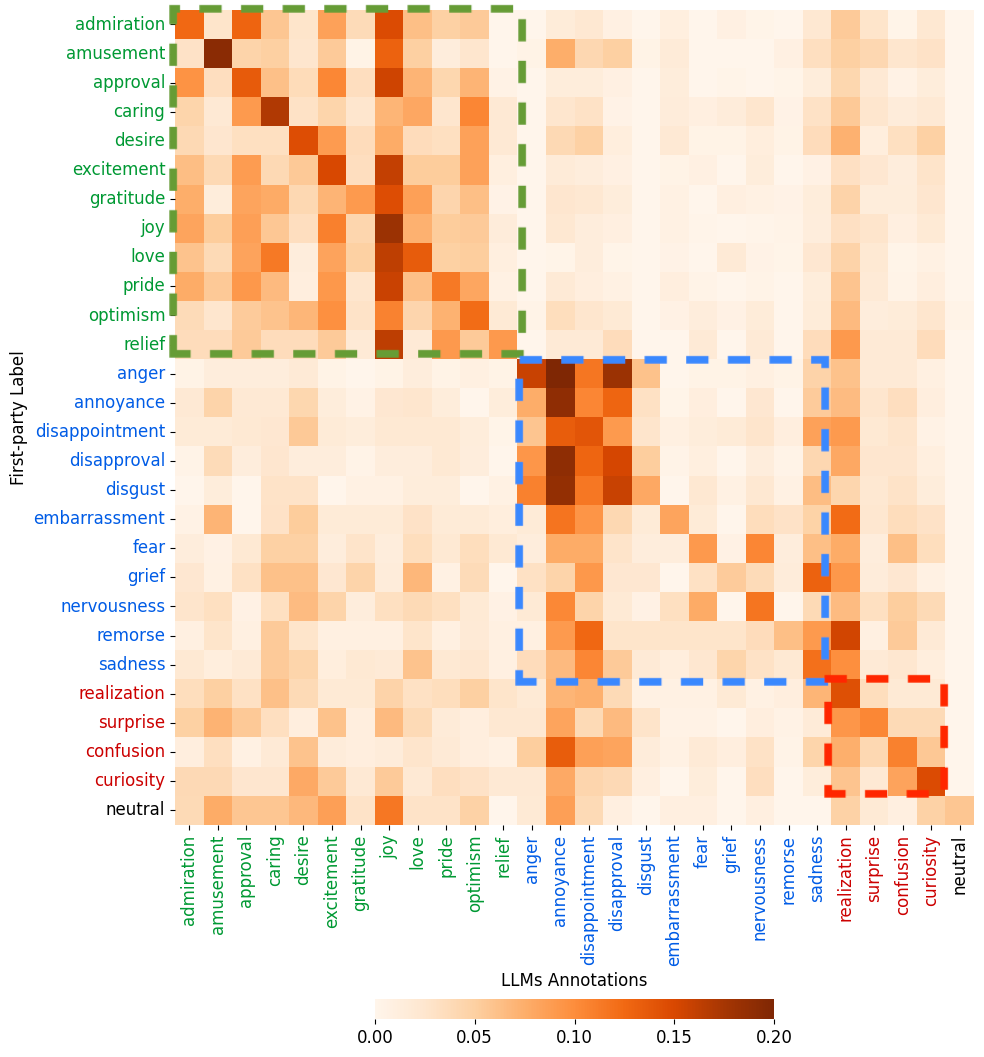 }
   \caption{Confusion matrix showing the alignment between LLMs' annotations and first-party emotion labels.}
   \label{fig:llm_confusion}
 \end{figure}
 
 \begin{figure}[t]
  \includegraphics[width=\columnwidth]{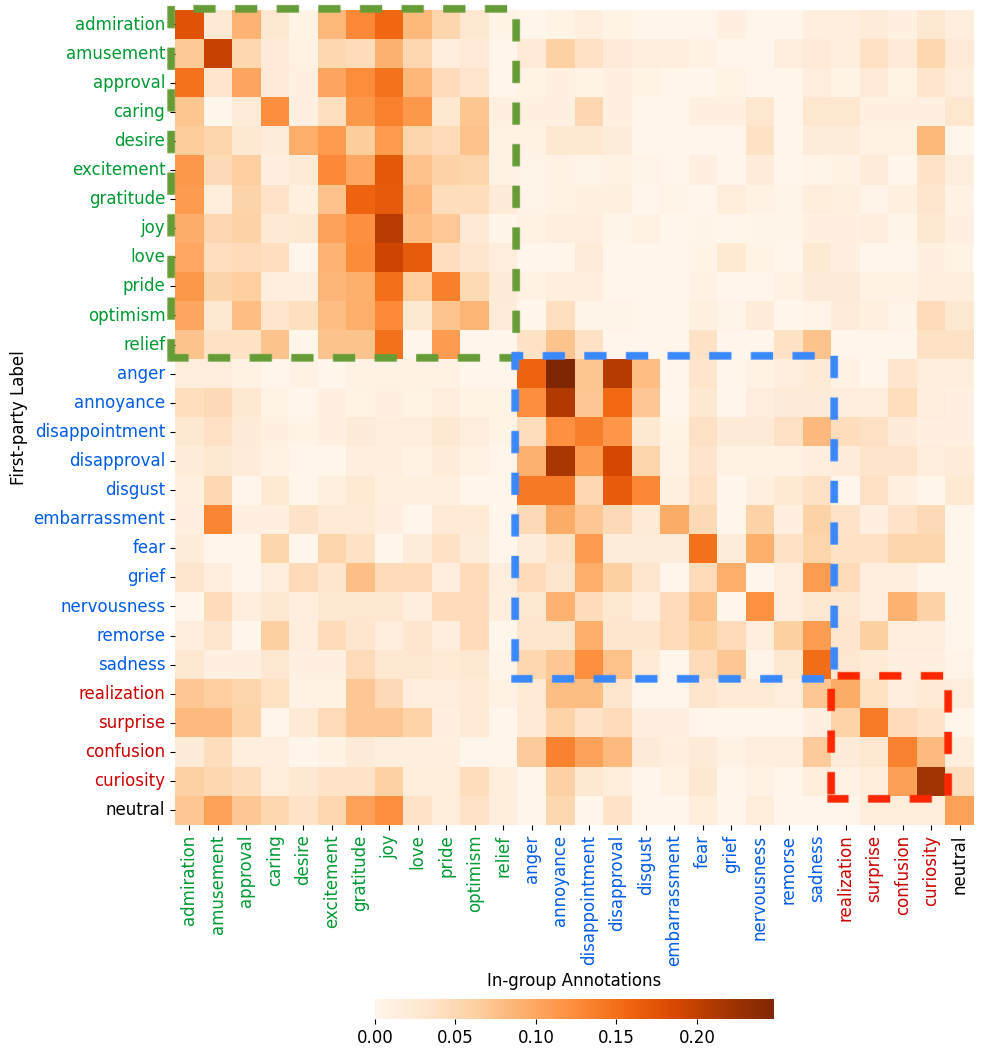 }
   \caption{Confusion matrix showing the alignment between in-group annotations and first-party emotion labels.}
   \label{fig:in_group_confusion}
 \end{figure}
 
\section{Agreement Among Third-party Annotators}
\label{agreement}

Figure \ref{fig:inter_rater} presents the interrater agreement amongst human annotators and amongst LLMs, evaluated by interrater correlation. LLMs generally exhibit higher and more consistent interrater agreement acorss emotions compared to human annotators. LLMs exhibit the highest agreement when labeling joy, love, and annoyance, while the lowest agreement is observed for neutral, confusion, and approval. In contrast, grief, sadness, and fear yield the highest interrater correlations among human annotators,  whereas desire, realization, and relief yield the lowest. 

Human annotators demonstrate higher agreement on negative emotions, while LLMs show higher agreement on positive emotions. Both human annotators and LLMs exhibit higher interrater agreement for emotions that are typically expressed with distinct textual cues and lower agreement for more nuanced and context-dependent emotions such as approval, relief, and realization. These emotions may be harder to infer externally due to their reliance on situational or implicit contextual cues.

We also computed the Cohen's Kappa by randomly sampling two annotations for each post and calculating the agreement between these two sets of annotations. We computed the correlation between Cohen's kappa values and interrater correlation and found that Cohen's kappa and interrater correlation are highly correlated 
(LLMs: Pearson $r = 0.83$, $p = 6.57 \times 10^{-8}$; 
Human annotators: Pearson $r = 0.71$, $p = 2.07 \times 10^{-5}$ \text{ for out-group}). This strong correlation suggests that both metrics capture similar trends in annotator agreement.

\section{First- and Third-Party Alignment}
\label{alignment_appendix}

Figure \ref{fig:cohenKappa} presents the alignment between first-party labels and third-party annotations, evaluated by Cohen's Kappa. The levels of alignment between third-party annotations and first-party labels vary across different emotions. While LLMs demonstrate more consistent alignment across emotions, human annotators show greater variation, which potentially reflect more subjective interpretation. Alignment is higher for emotions such as anger, love, and sadness for both human and LLM annotators, suggesting these emotions are more explicitly expressed and consistently recognized. Realization, relief, and approval yield the lowest alignment.

In addition, for each emotion, we calculated the percentage of posts where: (1) all annotators reached a decision; (2) a majority—but not all—of the annotators reached a decision; and (3) annotators failed to reach a decision due to ties. These percentages are reported in Table \ref{tab:A1}. For human annotations, on average, 27\% of posts for a given emotion received a majority (but not unanimous) decision, 70\% received a unanimous decision, and 3\% resulted in ties. For LLM annotations, on average, 16\% of posts yielded a majority decision and 84\% a unanimous decision, with no ties occurring (as each post is annotated by 5 LLMs). The large proportion of posts where disagreement occurs, which aligns with the relatively low inter-rater agreement among third-party annotators, further highlights the highly subjective nature of emotion recognition.

\begin{figure}[t]
\includegraphics[width=\columnwidth]{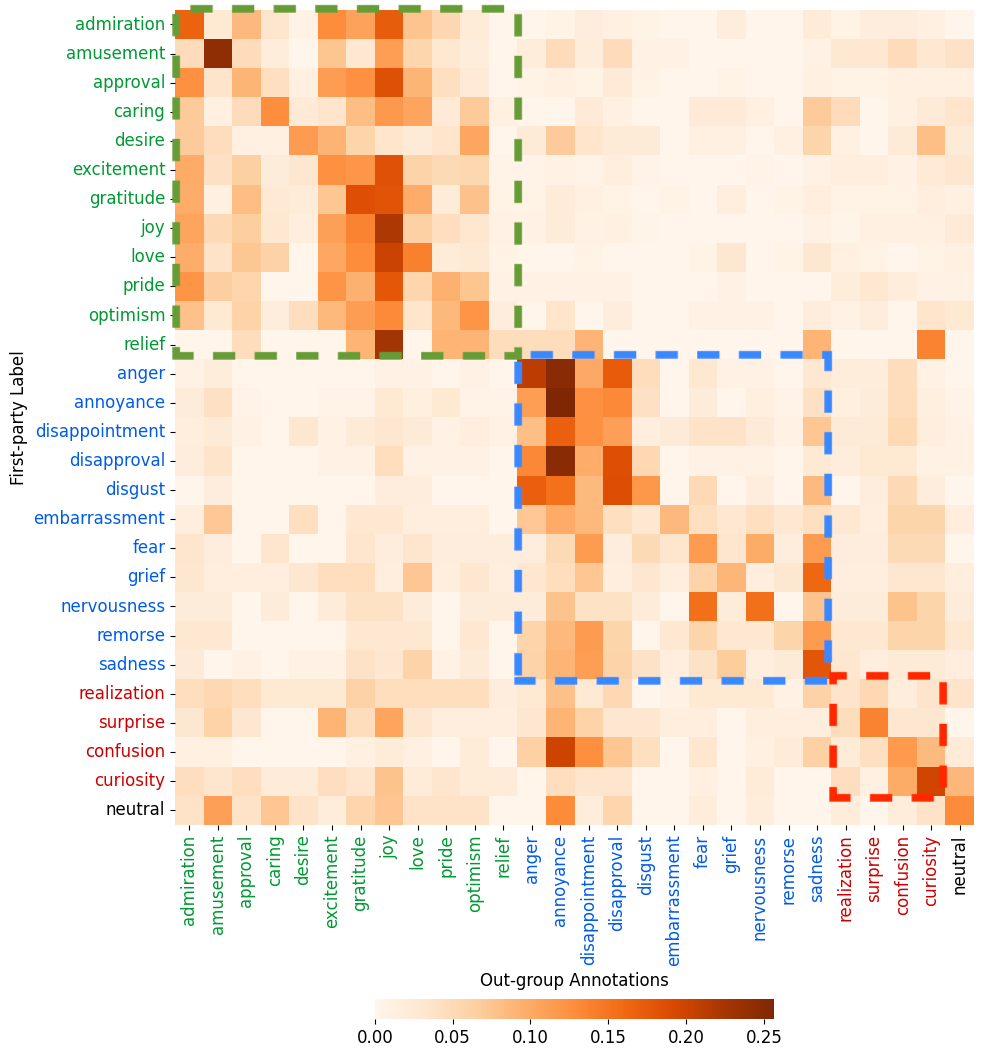 }
\caption{Confusion matrix showing the alignment between out-group annotations and first-party emotion labels.}
\label{fig:out_group_confusion}
\vspace{-0.3in}
\end{figure}

We further examined the confusion matrices (Figures \ref{fig:llm_confusion}, \ref{fig:in_group_confusion}, \ref{fig:out_group_confusion}) to understand the patterns of alignment and misalignment between first-party labels and third-party annotations. As shown in the figures, annotations frequently align with emotions within the same broad category—for example, positive emotions (e.g., joy, admiration, excitement) and negative emotions (e.g., anger, sadness, disappointment) tend to be more frequently confused with each other, as highlighted by the green and blue clusters. However, emotions that are more ambiguous or context-dependent, such as confusion, surprise, and realization, exhibit greater divergence between first-party and third-party labels, as seen in the red-outlined cluster. Additionally, LLMs show more consistent annotation patterns across emotions, whereas human annotators display greater variability, reflecting possible subjectivity in emotion interpretation.

\begin{table}[t]
\small
\centering
\begin{tabular}{lccc}
    \hline
    \textbf{Emotion} & \makecell{\textbf{Unanimous} \\ \textbf{Agreement}} & \makecell{\textbf{Majority} \\ \textbf{Agreement}} & \textbf{Ties} \\
    \hline
    Anger & 0.76 & 0.22 & 0.02 \\
    Love & 0.66 & 0.31 & 0.04 \\
    Sadness & 0.79 & 0.19 & 0.02 \\
    Joy & 0.53 & 0.41 & 0.06 \\
    Gratitude & 0.56 & 0.37 & 0.06 \\
    Amusement & 0.57 & 0.39 & 0.04 \\
    Annoyance & 0.60 & 0.32 & 0.08 \\
    Nervousness & 0.82 & 0.17 & 0.01 \\
    Disgust & 0.80 & 0.19 & 0.01 \\
    Curiosity & 0.77 & 0.20 & 0.02 \\
    Admiration & 0.51 & 0.43 & 0.07 \\
    Surprise & 0.73 & 0.24 & 0.03 \\
    Fear & 0.83 & 0.16 & 0.01 \\
    Disappointment & 0.62 & 0.33 & 0.05 \\
    Caring & 0.68 & 0.29 & 0.02 \\
    Grief & 0.87 & 0.12 & 0.01 \\
    Disapproval & 0.66 & 0.26 & 0.08 \\
    Pride & 0.65 & 0.30 & 0.05 \\
    Confusion & 0.75 & 0.23 & 0.03 \\
    Embarrassment & 0.86 & 0.13 & 0.01 \\
    Excitement & 0.61 & 0.32 & 0.07 \\
    No emotion & 0.71 & 0.28 & 0.02 \\
    Desire & 0.71 & 0.26 & 0.02 \\
    Optimism & 0.62 & 0.34 & 0.04 \\
    Realization & 0.65 & 0.32 & 0.03 \\
    Remorse & 0.87 & 0.12 & 0.01 \\
    Approval & 0.54 & 0.40 & 0.06 \\
    Relief & 0.78 & 0.21 & 0.01 \\
    \hline
\end{tabular}
\caption{Alignment between third-party annotations with first-party labels, evaluated by Cohen's kappa.}
\vspace{-0.2in}
\label{tab:A1}
\end{table}

\begin{figure}[t]
\includegraphics[width=\columnwidth]{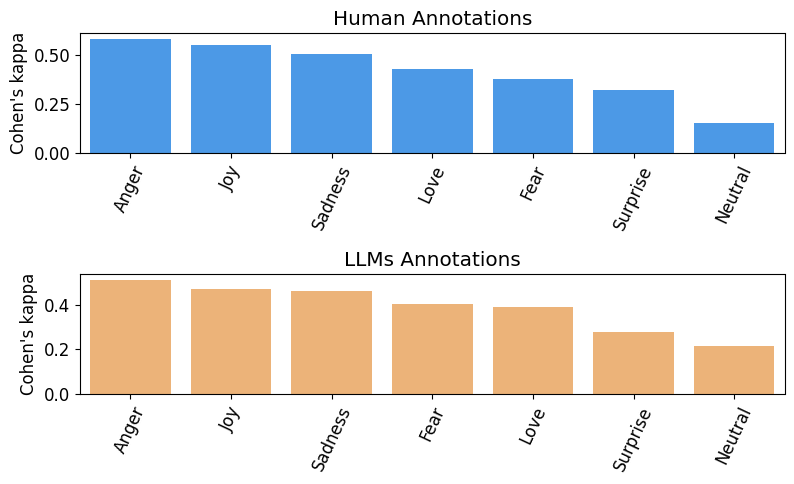}
\caption{Alignment between third-party annotations with first-party labels, on coarse-grained emotions, evaluated using Cohen Kappa. }
   \label{fig:human_llm_kappa_agg}
 \end{figure}

\section{Alignment Between Third-party Annotations and First-party Labels on Coarse-Grained Emotions}
\label{anaysis on aggregated emotion taxonomy}

\subsection{Coarse-Grained Emotion Taxonomy}
 To evaluate the alignment between third-party annotations and first-party labels at a higher level, we group the 28 emotion categories into seven groups: joy, love, anger, surprise, fear, sadness, and neutral. Since there is no universally agreed-upon set of basic emotions, we construct our grouping based the basic emotion categories proposed by \citet{ekman1992argument} (anger, disgust, fear, joy, neutral, sadness, surprise) and \citet{shaver1987emotion} (anger, love, fear, joy, sadness, and surprise).
\subsection{Third-party Annotators' Performance}
\label{third-party-annotation-performance}
We analyze the alignment between third-party annotations and first-party labels after mapping the original 28 emotion categories into 7 broader emotion groups. By aggregating annotations, we examine whether third-party annotators, while struggling to capture fine-grained emotions, align more closely with first-party labels when emotions are categorized at a higher level. For instance, annotations such as 'Admiration' and 'Approval' are grouped under 'Joy'. To assess alignment, similar to \ref{aligment}, we computed Cohen's kappa, F1, recall, and precision. Figure \ref{fig:human_llm_kappa_agg} presents Cohen's kappa scores for each emotion for human and LLM annotations. For human annotations, Cohen's kappa scores fall within the range of 0.15 - 0.6. For LLM annotations, Cohen's kappa scores fall within the range of 0.2 - 0.5.

\begin{figure}[h]
    \vspace{-0.1in}
  \includegraphics[width=\columnwidth]{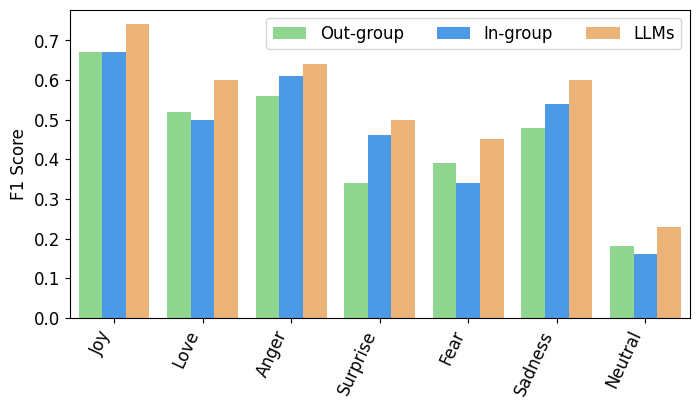}
  \vspace{-0.15in}
   \caption{Alignment between third-party annotations with first-party labels, on coarse-grained emotions, evaluated using F1 score}
   \label{fig:humn_llm_f1_agg}
 \end{figure}

The macro-average scores for LLM annotations are as follows: (F1: 0.54, Recall: 0.61, Precision: 0.57). For out-group human annotators, the scores are (F1: 0.45, Recall: 0.39, Precision: 0.55), while for in-group human annotators, they are (F1: 0.47, Recall: 0.41, Precision: 0.55). Figure \ref{fig:humn_llm_f1_agg} presents the F1 scores achieved by third-party human annotators (in-group, out-group) and LLMs across different emotions. These results suggest that third-party annotators struggle to identify the emotion expressed by the first party even at a higher level, indicating that this misalignment is not solely due to mistakenly interpreting similar emotions within the same broader category, but also stems from selecting emotions that are substantially different from those expressed by the first party. This highlights the inherent limitations in third-party annotations.

 \subsection{In-group Annotators VS. Out-group Annotators}
Similar to \ref{In-Group VS. Out-group}, we explored the impact of demographic similarity between first-party and third-party annotators on the alignment between first-party labels and third-party annotations when emotions are grouped into higher-level categories. We follow the same procedure as in Section \ref{In-Group VS. Out-group}. 

\noindent\textbf{Post-level comparsion} We applied the Wilcoxon signed-rank test to compare the F1, recall, and precision scores obtained by in-group majority-voted labels and out-group majority-voted labels for each post. The comparison is based on 93\% of posts. The Wilcoxon signed-rank test results indicate that in-group and out-group annotators perform similarly in terms of F1-score
(\(\text{Median}_{\text{in-group}} = 0.67\), \(\text{Median}_{\text{out-group}} = 0.67\), \(p = 0.05\)), 
with the difference being marginally significant. Precision also shows no significant difference 
(\(\text{Median}_{\text{in-group}} = 0.58\), \(\text{Median}_{\text{out-group}} = 0.57\), \(p = 0.602\)). 
However, in-group annotators achieve significantly higher recall than out-group annotators 
(\(\text{Median}_{\text{in-group}} = 1.00\), \(\text{Median}_{\text{out-group}} = 0.50\), \(p = 3 \times 10^{-3}\)), 
suggesting that they are better at capturing first-party expressed emotions at a broader level. 

\noindent\textbf{Annotator-level comparison} We adopted the same method presented in \ref{In-Group VS. Out-group} to test the impact of demographic similarity between third-party and first-party annotators at the annotator level. The results of the mixed linear models indicate that in-group annotators achieve significantly higher recall
(\(\text{Median}_{\text{in-group}} = 0.60\), \(\text{Median}_{\text{out-group}} = 0.56\), \(p = 0.006\)) 
and marginally higher F1-score (\(\text{Median}_{\text{in-group}} = 0.53\), \(\text{Median}_{\text{out-group}} = 0.51\), \(p = 0.04\)) 
compared to out-group annotators. However, no significant difference is observed in precision 
(\(\text{Median}_{\text{in-group}} = 0.54\), \(\text{Median}_{\text{out-group}} = 0.52\), \(p = 0.29\)).These findings suggest that demographic similarity continues to play a role in improving annotation quality, even when evaluating alignment between third-party annotations and first-party label at a coarse-grained level of emotion categorization. 

\section{LLM Prompts}
\label{prompt}
\begin{figure*}[h]
  \includegraphics[width=0.95\textwidth]{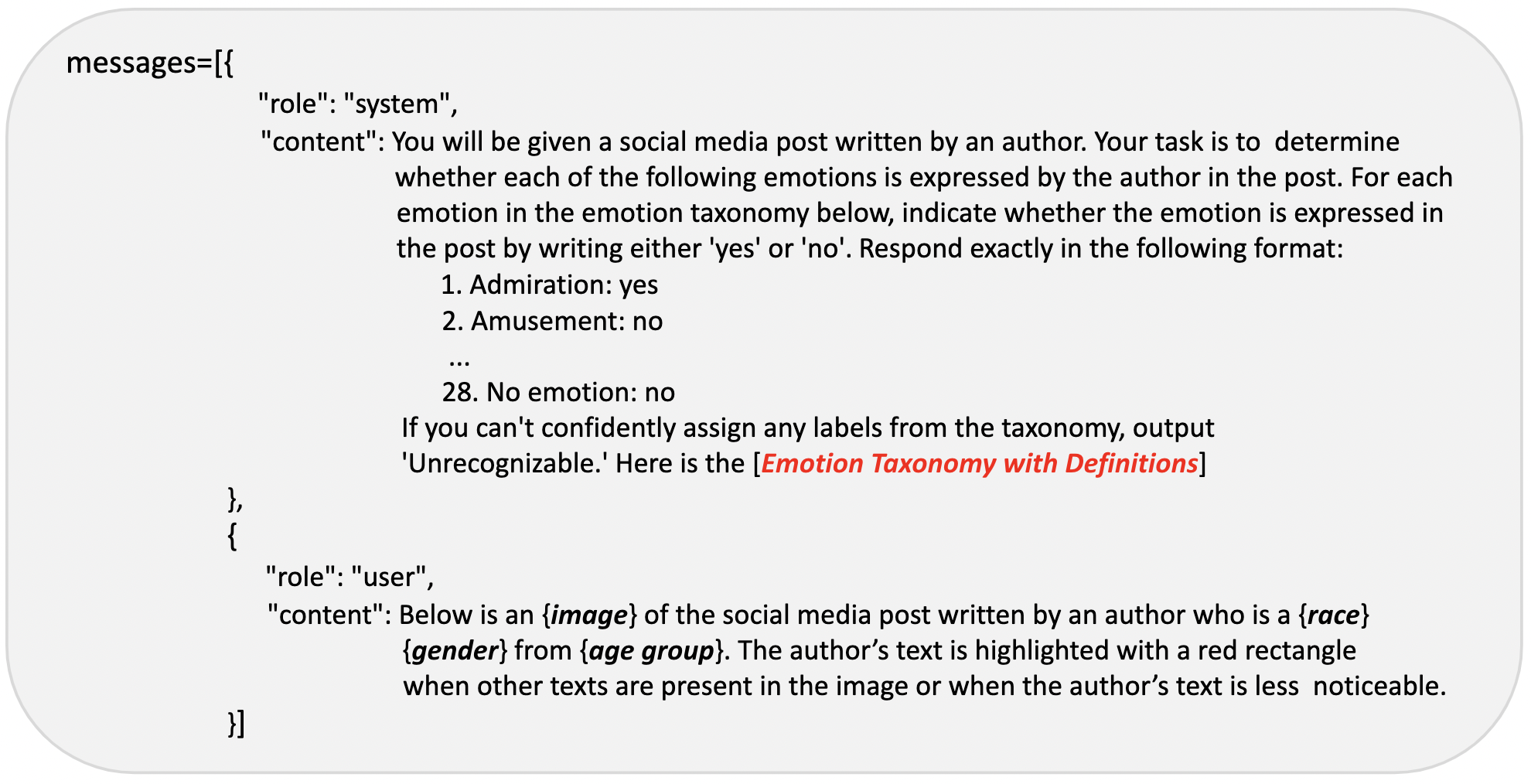}
   \caption{Prompt used to elicit LLM annotations.}
   \label{fig:promt1}
\end{figure*}

\begin{figure*}[t]
  \includegraphics[width=0.95\textwidth]{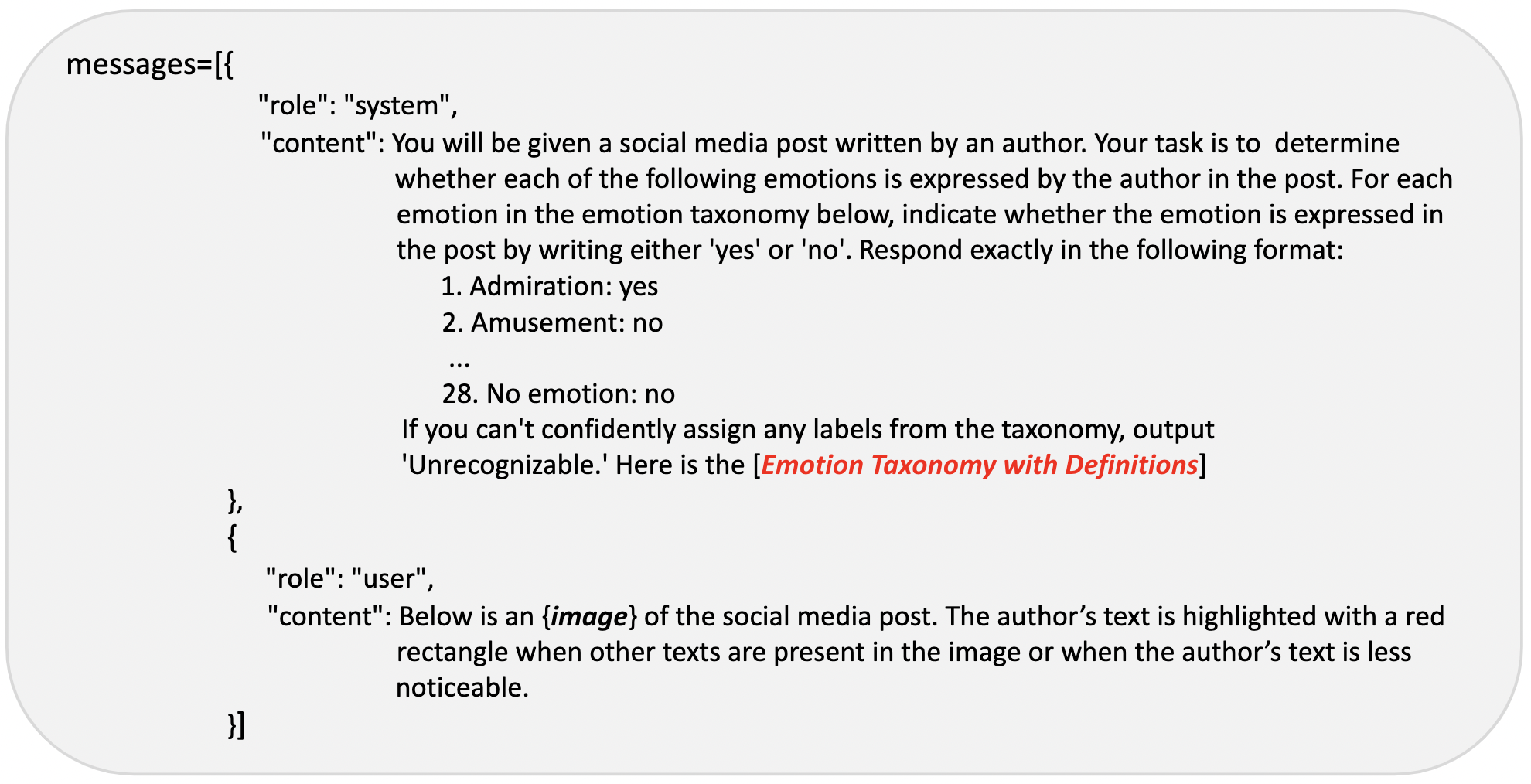}
   \caption{Prompt with first-party demographic information.}
   \label{fig:promt2}
 \end{figure*}

We introduce the LLM prompts for annotations, as presented in Figure \ref{fig:promt1} and \ref{fig:promt2}. We also inserted first-party age group, gender, and race into the prompt of Figure \ref{fig:promt1}. We applied the same prompt for all LLMs.

  \begin{figure*}[t]
   \includegraphics [width=0.9\textwidth]
   {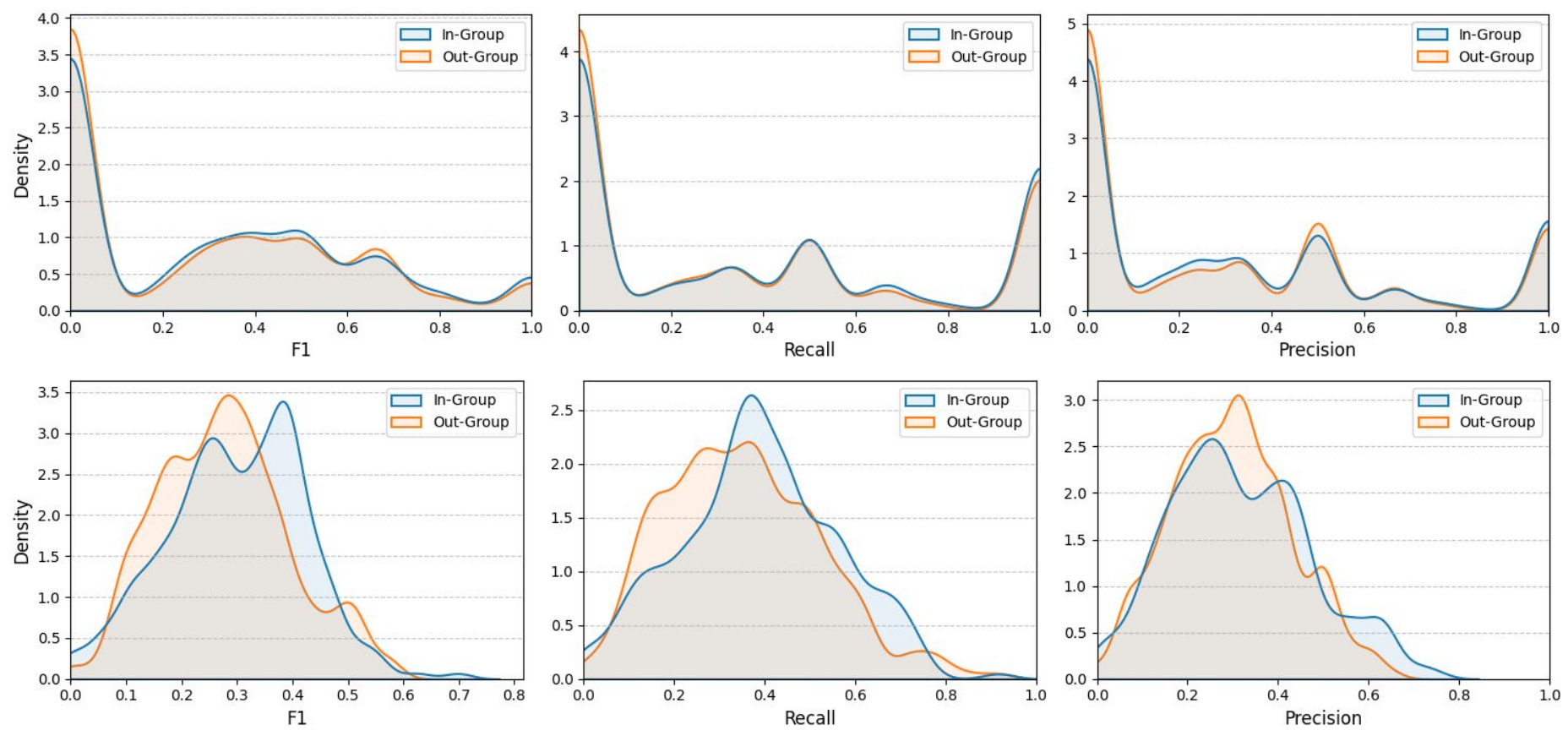}
    \caption{Density plots showing the distribution of F1 scores, recall for each post (first row). Density plots showing the distribution of averaged F1 scores, recall individual annotators per task (second row).}
    \label{fig:f1_dist}
  \end{figure*}

 \begin{figure*}[h]
  \includegraphics[width=\textwidth]{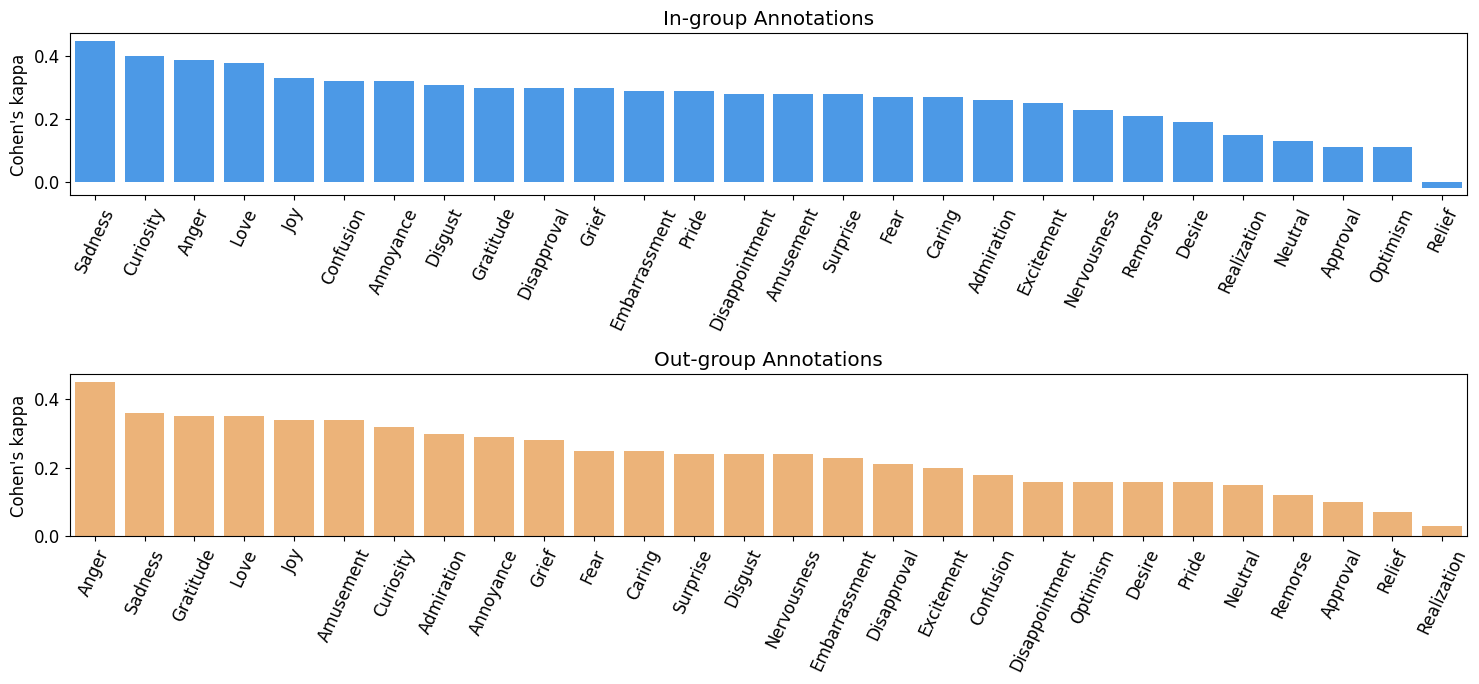 }
   \caption{Performance comparison of in-group annotators and out-group annotators by Cohen's kappa}
   \label{fig:cohenKappa_in_out}
 \end{figure*}

 \begin{figure*}[h]
  \includegraphics[width=\textwidth]{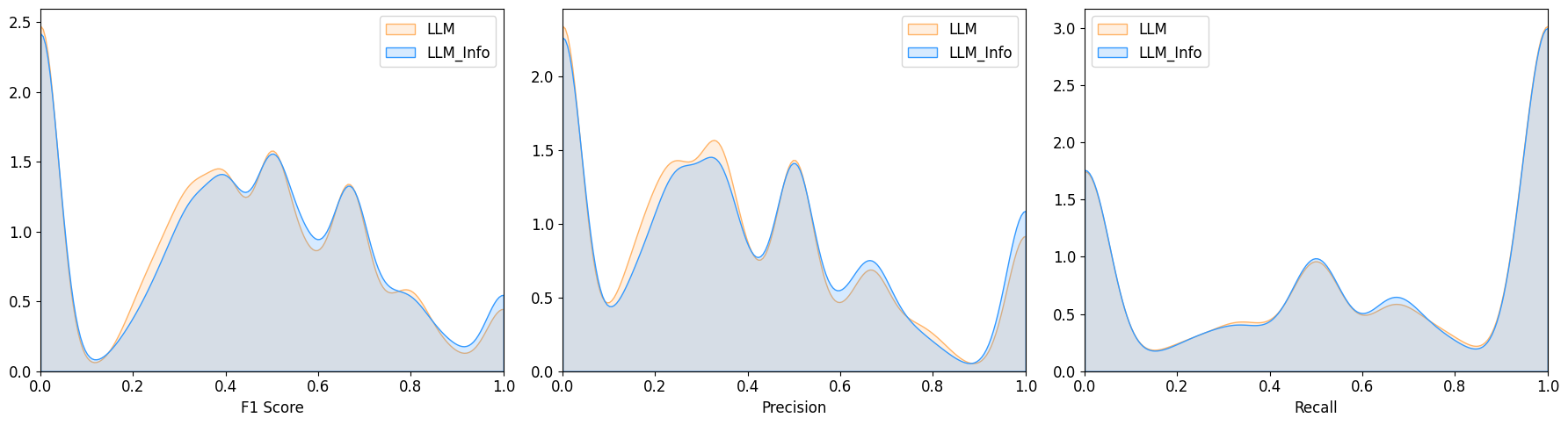 }
   \caption{Density plots showing the distribution of F1 scores, recall, and precision for each.}
   \label{fig:llm_llm_info}
 \end{figure*}

\end{document}